\documentclass{article}

\usepackage[main, final]{neurips_2026}

\usepackage[utf8]{inputenc} % allow utf-8 input
\usepackage[T1]{fontenc}    % use 8-bit T1 fonts
\usepackage{hyperref}       % hyperlinks
\usepackage{url}            % simple URL typesetting
\usepackage{booktabs}       % professional-quality tables
\usepackage{amsfonts}       % blackboard math symbols
\usepackage{nicefrac}       % compact symbols for 1/2, etc.
\usepackage{microtype}      % microtypography
\usepackage{xcolor}         % colors

\usepackage{microtype}
\usepackage{graphicx}
\usepackage{subcaption}

\usepackage{amsmath}
\usepackage{amssymb}
\usepackage{mathtools}
\usepackage{amsthm}

\newcommand{\xhdr}[1]{\vspace{2mm}\noindent{{\bf #1.}}}
\usepackage{amsthm}

\theoremstyle{plain}
\newtheorem{theorem}{Theorem}[section]

\theoremstyle{definition}
\newtheorem{definition}[theorem]{Definition}

\theoremstyle{remark}

\newcommand{\levelup}{\textsc{LevelUp}}
\newcommand{\leveldown}{\textsc{LevelDown}}
\newcommand{\random}{\textsc{Random}}

\title{Level Up: Defining and Exploiting Transitional Problems for Curriculum Learning}

\author{%
  Amogh Inamdar\thanks{Equal contribution} \\
  Department of Computer Science\\
  Columbia University\\
  \texttt{amogh.inamdar@columbia.edu} \\
  \And
  Zhenwei Tang\footnotemark[1]\\
  Department of Computer Science\\
  University of Toronto \\
  \texttt{josephtang@cs.toronto.edu} \\
  \And
  Ashton Anderson \\
  Department of Computer Science\\
  University of Toronto \\
  \texttt{ashton@cs.toronto.edu} \\
  \And
  Richard Zemel\\
  Department of Computer Science\\
  Columbia University \\
  \texttt{zemel@cs.columbia.edu} \\
}

\begin{document}

\maketitle

\begin{abstract}
  Curriculum learning---ordering training examples in a sequence to aid machine learning---takes inspiration from human learning, but has not gained widespread acceptance. Static strategies for scoring item difficulty rely on indirect proxy scores of varying quality and produce curricula that are not specific to the learner at hand. Dynamic approaches base difficulty estimates on gradient information, requiring considerable extra computation during training. We introduce a novel method for measuring the difficulty of individual problem instances that is calibrated to a series of models of increasing competence, and identify \emph{transitional problems} that are consistently easier as model ability increases. Applying this method to diverse model series constructed from sets of models that are readily available on many tasks, we find that training on a curriculum that \emph{levels up} from easier to harder transitional problems most efficiently improves a model to the next tier of competence. These problems induce a natural progression from easier to harder items, which outperforms other training strategies. By measuring difficulty directly relative to model competence, our method yields interpretable problems, learner-specific curricula, and a principled basis for step-by-step improvement.
\end{abstract}

\section{Introduction}\label{sec:intro}

Machine learning (ML) differs significantly from human learning in practice, both in terms of the learner and the learning method. Human learning is generally curriculum-driven, with a training diet of examples that increase in complexity as the learner improves in ability. On the other hand, an ML model optimized on data drawn i.i.d.\ from the training distribution (i.e., empirical risk minimization, or ERM) typically learns to perform a task at least as well as when optimized on the same examples presented in a structured manner. Curriculum learning \citep{elman1993learning,sanger2002neural,bengio2009curriculum}---the process of training ML models on a structured data regimen, as opposed to random samples---has shown benefits in some settings, particularly for models that are trained on noisy or limited data \citep{wu2020curricula, wang2023efficienttrain}, in models for sequential decision making such as in reinforcement learning \citep{tao2024reverse, zhao2022learning}, and in the post-training of foundation models for mathematical reasoning \citep{zeng2025glm}. However, several negative results for curriculum learning temper these successes. Curricula have shown no benefit in end performance over ERM on high-quality or very large multi-task datasets \citep{wu2020curricula}, with overparameterized networks \citep{mannelli2024tilting}, and even when training on data that emulates human learning \citep{warstadt-etal-2023-findings}. 

A primary consideration in curriculum learning is the definition of an appropriate measure of the difficulty of a training example. The structure of a curriculum for human learners depends not only on the relations between the concepts being learned, but also on the learning outcomes and experiences of past learners.
Difficulty measures for ML models typically rely on domain-specific and human-centric information that may not translate well to ML models (e.g., in BabyLM submissions \citep{oba2023babylm}), or on the evolving training dynamics of the learner (such as the gradient norm \citep{graves2017automated}), which can add considerable expense to training.
We observe that very few difficulty measures for curriculum learning on ML tasks leverage the diverse collection of existing models for that task. Analogous to the range of performance within and across human learners, these models vary in task proficiency due to \emph{inter-model} differences (size, architecture, complexity of training data, etc.) and \emph{intra-model differences} (task-specific context, input, time/compute resources, etc.).
Inspired by this observation, as well as the 
stratification of many human curricula into multiple `levels' (such as the scholastic grades in the K-12 education system), we ask the following questions: 
(1) Are there specific problems that consistently mark the boundary between the competence levels of an existing collection of ML models?
(2) Can we identify these specific problems at each level, and their common characteristics? 
(3) Does training on such problems provide an efficient way for the learner to make progress towards the next level (e.g., $4^\text{th}\rightarrow5^\text{th}$ grade in K-12), and inform a strategy for curriculum learning?

We answer these questions in the affirmative by defining and identifying \emph{transitional problems}---problems that exhibit a sharp transition in solvability across increasing competence levels. 
Given a series of models ordered by increasing performance, a transitional problem at a given level can be solved by all models at or above that level, but not by those below. 
These problems mark competence boundaries and yield an empirically grounded easy-to-hard partitioning of the training data distribution. We develop methods to characterize the levels of competence of an ML model, identify transitional problems, and build a curriculum inspired by human learning.
Through experiments across a variety of model series, we show that a {\levelup} curriculum on transitional problems leads to better learning outcomes than other data orderings (including i.i.d.) on both transitional problems and held-out test problems.
Finally, we examine the characteristics of transitional problems and find correlations with human difficulty measures, despite no explicit optimization for interpretability.

\section{Related Work}\label{sec:related}

\xhdr{Efficacy of Curriculum Learning} Despite the ubiquity of curricula in human learning, curriculum learning remains niche in machine learning practice, with multiple negative results in the literature. The BabyLM challenge \citep{oba2023babylm} for developmentally-inspired LLM pretraining saw a variety of domain-specific \citep{martinez2023climb, edman2023too, oba2023babylm}, model-dependent \citep{opper2023effect}, and student-teacher curricula \citep{chobey2023can, zhang2023baby} fail to beat the baseline. Other works show that curricula provide no benefit with large training budgets and clean data \citep{wu2020curricula} or over-parameterized models \citep{mannelli2024tilting}. Fine-grained studies of curriculum learning primarily show benefits in low-data or high-noise regimes \citep{wu2020curricula}. However, recent work has shown wins for curriculum learning in structured domains like reinforcement learning (for long-horizon tasks \citep{narvekar2020curriculum, patel2402curriculum, li2025causally, zhao2022learning} and faster convergence \citep{tao2024reverse}) and LLM post-training for instruction-following \citep{ge2025dynamic} and complex reasoning \citep{zeng2025glm, polu2022formal}. The training of large-scale foundation models is increasingly divided into pre-, mid-, and post-training phases \citep{ouyang2022training, wang2025octothinker, olmo20242}, with different training tasks, datasets, and learning objectives in each phase \citep{liu2024deepseek, zeng2025glm}. While these phased approaches operate at a coarser granularity than example-level curriculum learning, they reflect a similar principle of structuring the training process. At a fine-grained level, the GPT-3 and T5 LLMs had non-uniform mixing strategies \citep{brown2020language, raffel2020exploring} in pretraining, and recent work shows additional promising results for curricula in pretraining \citep{zhang2026beyond, liu2026languagemodelslearnwhen}.

\xhdr{Static and Dynamic Curricula} Curriculum learning methods are typically designed as \emph{static} strategies for ordering and pacing the training data input to a model \citep{bengio2009curriculum, weinshall2018curriculum, jiang2020characterizing}, focusing on long-term improvement. The initial cost of curriculum design is amortized over a long training process or by applying to multiple learners, making static curricula suitable for large-scale training such as in structured pre- and post-training \citep{brown2020language, wake2024yi} and length generalization \citep{liu2024deepseek} in LLMs. An alternative approach is to \emph{dynamically} adapt the data diet based on a model's learning performance during training \citep{bellemare2016unifying, graves2017automated, zhang2021flexmatch}. While adaptivity can lead to a better learning signal in some settings, dynamic curricula require periodic re-evaluation and recomputation of difficulty during every training run. While both dynamic and static approaches have their advantages, in this work, we focus on developing and evaluating a \emph{static} curriculum. We present additional details on these strategies in Appendix \ref{apx:more-related}.

% Curriculum learning methods can broadly be categorized as either \emph{dynamic} or \emph{static}. The former leverage training dynamics to adapt the input to a model over the learning process \citep{bellemare2016unifying, graves2017automated, zhang2021flexmatch}, even making per-sample choices in the limiting case \citep{cohn1994improving}. While dynamic curricula may adapt to fit model needs, they training-time complexity by periodically recomputing difficulty, and are thus less popular for large-scale model training at the time of writing. In contrast, \emph{Static} curricula establish the strategy and pacing of a curriculum prior to the learning process \citep{bengio2009curriculum, weinshall2018curriculum, jiang2020characterizing}. Static methods focus on efficient long-term improvement by structuring the full training process prior to learning. The initial cost of curriculum design can be amortized over a long training process or by applying to multiple learners, making static curricula suitable for large-scale training such as in structured pre- and post-training \citep{brown2020language, wake2024yi} and length generalization \citep{liu2024deepseek} in LLMs. While both dynamic and static approaches have their advantages, in this work, we focus on developing and evaluating a \emph{static} curriculum.

\xhdr{Difficulty, Ordering, and Pacing} The design of a static curriculum typically involves three key decisions \citep{wu2020curricula}. The first (and perhaps defining) choice is that of selecting an appropriate \emph{item difficulty strategy}, which is a measure of the utility of a training example for learning. Static curricula typically focus on \emph{model-independent} difficulty measures, such as the quality of an image \citep{wang2023efficienttrain, sheybani2023curriculum}, complexity of a reasoning problem \citep{polu2022formal}, or the size and quality of textual data \citep{zhang2018empirical, warstadt-etal-2023-findings}. These leverage the innate complexity of the training data, but do not account for the specific characteristics of a learning model. While model-derived approaches are rare, we highlight two key works in this area. The first scores item difficulty by training an SVM \citep{cortes1995support} on the representations of teacher model for image classification \citep{weinshall2018curriculum, hacohen2019power}. This approach relies on an external model of the performance and dynamics of a model (the SVM), and may not easily generalize to other domains. The second aggregates the performance of models trained on diverse pretraining sets as the \emph{consistency score} (C-score) of an example \citep{jiang2020characterizing}. The C-score robustly evaluates learning difficulty across multiple models, but does not take the strength of each model into account when aggregating difficulty. In this work, we treat this group as a \emph{series} of models of progressing strength, and develop a method to identify the \emph{transitional problems} that characterize the relative difference in competence between such models. We show that leveraging such a series produces a measure of difficulty that leads to better outcomes with curriculum learning than ordering by the C-score. 

% The latter produces a robust metric, but does not account for any misalignment between the pretraining and target tasks or difference in teacher utility. Thus, a training problem that is only consistently solved by strong teachers may be scored as equally difficult to a problem that happens to be solved by shortcut learning or memorization. In this work, we develop a \emph{task-specific measure of difficulty that accounts for the strength of each teacher}.

Once a strategy for item difficulty is set, the \emph{ordering} of the samples by difficulty (easy-to-hard, hard-to-easy, random) as well as the \emph{pacing}
(training effort per difficulty level) must be set. However, finding the optimal pacing (constant, linear, exponential, etc.) requires considerable effort, and in some cases provides only a marginal benefit \citep{wu2020curricula}. Thus, in this work, we set a simple \emph{constant} pacing strategy for every curriculum, with each difficulty level receiving equal attention. This ensures that performance differs only based on the difficulty measure and ordering.

\xhdr{Human Learning} Our strategy for `leveling up' ML models with transitional problems is motivated by the structure of curriculum-driven human learning. A learning regimen for humans typically emphasizes sub-tasks that are just out of reach of a learner's current capabilities. In psychology, the technique of \emph{scaffolding} \citep{wood1976role} is used to teach infants new skills (such as object detection and manipulation) by focusing on tasks in their \emph{Zone of Proximal Development} (ZPD) \citep{vygotsky1978mind}, i.e., tasks that can be accomplished with some teacher assistance. This is analogous to the concept of training on the transitional problems of the next competence level. Neural networks trained on egocentric videos from infants perform best when trained on a developmental (young-to-old) ordering \citep{sheybani2023curriculum}. Even beyond infancy, ordering concepts by complexity is the standard in human learning, even for physical tasks such as motor skills \citep{sungeelee2024interactive}.
\section{Transitional Problems}\label{sec:method}

\begin{figure}[h!]
    \centering
    \includegraphics[width=\linewidth]{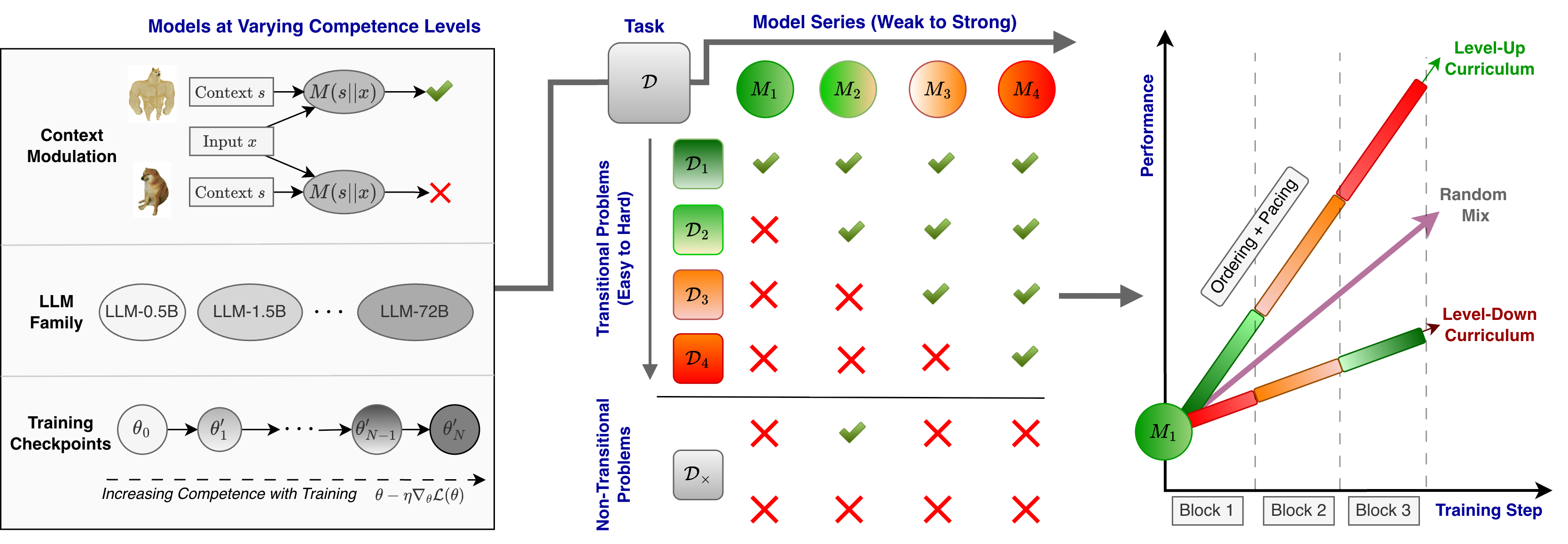}
    \caption{Transitional Problems at a level $i$ can only and consistently be solved by models at a competence level $j\ge i$. We find that with modest budgets, training on an ascending-level curriculum on transitional problems produces models that perform better than those trained with other strategies.} \label{fig:transitional_fig1}
    % \vspace{-1cm}
\end{figure}

\subsection{Models at Varying Levels of Competence}\label{method:model-series}

Given a collection of ML models $\mathbb{M}$ on a task with data distribution $\mathcal{D}$, our goal is to design a curriculum for the rapid improvement of a base model $M_0$. We begin by selecting a subset of these models that form \emph{model series} of monotonically increasing \emph{strength} (performance) $s$.

\begin{definition}[Model Series]
\label{def:model_series}
Let $\mathcal{M}\subseteq\mathbb{M} = \{M_0, M_1, \ldots, M_n\}$ be a finite set of $n+1$ models. $\mathcal{M}$ forms a \emph{model series} w.r.t. data distribution $\mathcal{D}$ and performance measure $s: \mathcal{M} \rightarrow \mathbb{R}^+$ iff:
\begin{itemize}
    \item Each model has a unique strength: $\forall i, j \in \{1, \ldots, n\}, i \neq j \implies s(M_i) \neq s(M_j)$; and
    \item The models are indexed by increasing strength: $s(M_i) < s(M_j)$ for all $0 \leq i < j \leq n$
\end{itemize}
\end{definition}

The variation in performance of models within such a series may be due to \emph{inter-model} differences or \emph{intra-model} differences (Section \ref{sec:intro}). For example, a family of Large Language Models (LLMs) shows an inter-model progression in size (from 0.5--100+ billion parameters)---and correspondingly, competence---on a broad swathe of ML tasks. Another example is the series of \emph{training checkpoints} collected for error recovery and to study training dynamics, which form a natural progression in strength as their parameters adapt to a task. In contrast, a single LLM can exhibit intra-model variation through \emph{few-shot in-context learning} \citep{brown2020language}, which updates the context with task-specific examples to improve model performance. Models like the Maia-2 \cite{tang2024maia2} can adapt their performance based on strength-modulating contextual inputs. We depict these groups of models in Figure \ref{fig:transitional_fig1} (left).

\subsection{Defining Transitional Problems}\label{method-transitional}

Given a series of models that differ in performance on a learning task, a natural definition for the learning difficulty of an example is then the 
level of the weakest model that is `correct' on that example.\footnote{We focus on problem solving (with binary correctness) here, but appropriate thresholding can generalize this approach.} However, a weak model may get a hard problem right due to shortcut learning, or just by chance. Thus, we add a \emph{monotonicity} constraint to this definition: for a given problem $p$ to be at level $i$, any model that is below level $i$ must fail to answer $p$, and any at or above level $i$ must answer $p$ correctly. We designate examples that satisfy this constraint as the \emph{transitional problems} (Figure~\ref{fig:transitional_fig1}) corresponding to that model series, to indicate the smooth transition in easiness as models grow in competence. 
In concrete terms, we define the \emph{transitional problems} $\mathcal{T}\subseteq\mathcal{D}$ as follows.

\begin{definition}[Transitional Problem]
\label{def:transitional_problem}
Let $\phi_p(M_i) = 1$ if model $M_i$ correctly solves problem $p$, and $0$ otherwise. A problem $p$ is transitional w.r.t. $\mathcal{M}$ at level $1\le k \le n$ if:
\begin{itemize}
    \item $\forall i \in [k-1]: \phi_p(M_i) = 0 \land \forall i \in [k, n]: \phi_p(M_i) = 1$
\end{itemize}
We call $\tau_p = k$ the \emph{transition point} and $M_{k}$ the \emph{transition model} for problem $p$.
\end{definition}

\begin{definition}[Transitional Problems at $\tau$]
\label{def:transitional_problems_tau}
For a data distribution $\mathcal{D}$ and a transition point $\tau \in [n]$, the transitional data distribution $D_\tau$ at point $\tau$ is defined as:
\begin{equation}
\mathcal{D}_{\tau} = \{p\in\mathcal{D}:\phi_p(M_i)_{i<\tau}=0\land \phi_p(M_i)_{i\ge\tau}=1\}
\end{equation}
\end{definition}

Restricting a dataset to transitional problems produces a \emph{partially ordered subset}---problems at a given level are equivalent, but are strictly harder (easier) than ones at a lower (higher) level. 
Notably, this difficulty is with respect to the model series, not based on a human-centric metric. 
We construct curricula on three \emph{orderings} of transitional problems at constant \emph{pacing}. The {\levelup} curriculum trains a model $M_i$ at level $i$ on transitional problems at levels $T_{i+1}\rightarrow T_{i+2}\rightarrow\cdots\rightarrow T_n$. {\leveldown} reverses this order, training $M_i$ on $T_n\rightarrow\cdots\rightarrow T_{i+1}$. 
Finally, {\random} trains on a uniformly random sample from the set $T=\bigcup_{j=i+1}^n T_j$. In following sections, we show that transitional problems represent an easy-to-hard ordering of training data---both in terms of learning difficulty and along some human-interpretable features---through experiments with these and baseline curricula across a range of inter-model and intra-model series that are typical in ML settings. 
\section{Inter-Model Transitional Problems: Math}
\label{sec:experiments-math}
\label{expt:math-curr}
\label{subsec:expt-math-family}

As outlined in Section \ref{method:model-series}, a model series can be constructed in many ways. In this section, we primarily study transitional problems derived from various types of \emph{inter-model} series. In many domains, practitioners develop a series of models that trade off raw performance for other factors (efficiency, cost, runtime, etc.). This is especially the case in the vision and language domains, particularly on reasoning tasks. Models that can reason well in natural language (e.g., LLMs) are typically large and expensive, leading to the development of smaller models (SLMs with $\le 4$ billion parameters). LLM reasoning can be improved permanently with supervised training or temporarily in-context learning \cite{brown2020language}. Given this diversity, we select \emph{mathematical reasoning with SLMs} as our first testbed for transitional problems. We conduct a comprehensive evaluation of curriculum learning across 3 types of model series---distillation, checkpointing, few-shot adaptation---and find that transitional problems are an effective instrument for curriculum learning. We also show that transitional problems on one dataset can be used to find partition problems into levels of difficulty on a related dataset, and that this \emph{transfer} can lead to improvements with curricula on the latter dataset. We also find patterns in difficulty that are absent in the closest related grouping: the sets of \emph{all problems solved by a model at a given level of competence}.

\xhdr{Dataset and Models} Our choice of models and dataset is determined by two factors. First, the weaker models in the series must show a low base performance on the dataset, and improve considerably with training. This enables us to study the effects of curriculum learning in settings where the model performance is not too high (i.e., saturated) or too low to be able to observe differences. Second, the dataset must be reasonably interpretable and admit curricula on domain-specific difficulty measures. Based on these criteria, we choose the Qwen2.5 family of LLMs---with $\{0.5, 1.5, 7, 14, 32\}$ billion parameters---as our series \citep{qwen2025qwen25technicalreport}, and the GSM8k dataset \cite{cobbe2021training} as our reasoning task. The smaller Qwen2.5 models show a low 0-shot performance on GSM8k, but improve considerably with training. The larger ones perform well, making this combination ideal for our experiments. Unlike datasets for competition-level math \citep{hendrycks2021measuring} or formal reasoning \citep{zheng2021minif2f}, GSM8k solutions are well-formatted (with 2--10 annotated steps of grade-school arithmetic each) and are easy to interpret, allowing for comparisons with domain-specific and dataset-dependent curricula. 

\xhdr{Baselines} Aside from training on randomly sampled problems (`IID'), we also benchmark our transitional curricula on two dataset-specific measures of difficulty: the \emph{length of the example} (longer problems are more difficult to intepret and solve) and the \emph{number of reasoning steps in the solution} (annotated by GSM8k). We also evaluate a curriculum based on the C-score \citep{jiang2020characterizing}, which ranks the easiness of a problem as the number of models in the series that correctly solve it. In each experiment, we train the model on each curriculum that is implementable on the given dataset, and evaluate its average accuracy on the held-out test split of that dataset. We evaluate the accuracy on a given problem as the average success rate over 8 attempts per problem (i.e., \emph{Avg@8}) to account for the variance in model generations. Additional details are presented in Appendix \ref{apx:math_details}.

\begin{figure}
    \centering
    \begin{minipage}{0.48\textwidth}
        \includegraphics[width=\linewidth]{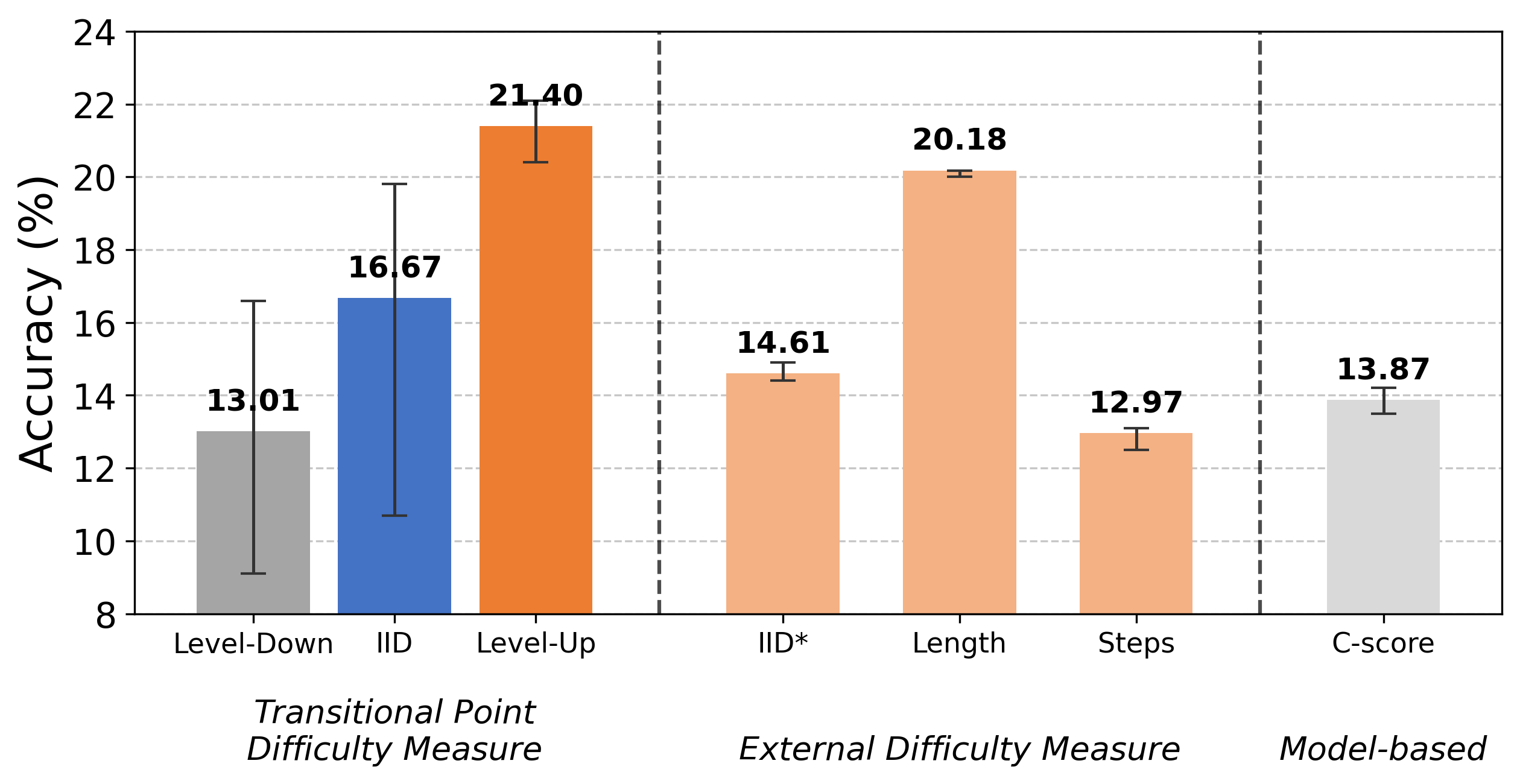}
        \caption{\textbf{Math:} Distillation from the Qwen2.5 model family into Qwen2.5-0.5B-Base. {\levelup}, corresponding to progressive distillation, outperforms other orderings and baselines.}\label{fig:math-family-0.5b}
    \end{minipage}%
    \hfill
    \begin{minipage}{0.48\textwidth}
        \centering
        \includegraphics[width=\linewidth]{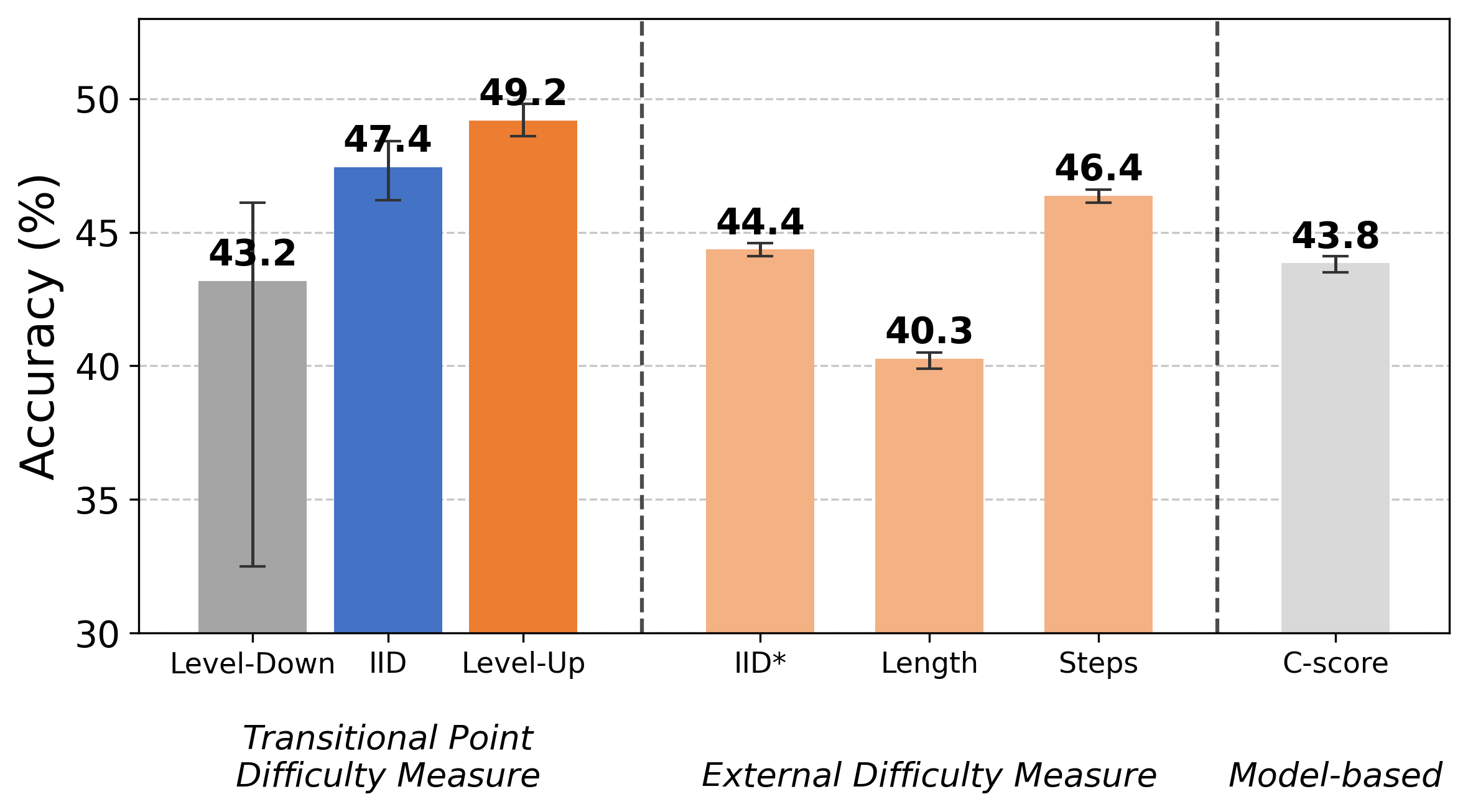}
        \caption{\textbf{Math:} Curricula on transitional problems collected from Qwen2.5-1.5B training checkpoints on GSM8k. {\levelup} outperforms {\random}, {\leveldown}, and all baselines.}\label{fig:math-curr-1.5b}
    \end{minipage}%
% \vspace{-0.5cm}
\end{figure}

\xhdr{Series A: Distillation} SLMs are typically trained by distilling information---usually output logits---from their larger counterparts \cite{hinton2015distilling, xu2024survey}. Prior work shows improved results from \emph{progressive distillation} \citep{mirzadeh2020improved, panigrahi2024progressive}, where knowledge is distilled from increasingly stronger models. This is a natural setting for our method, which also leverages a model series ordered by strength. Crucially, our method does not require the full outputs of a model, only correctness, enabling learning from closed-weight LLMs like ChatGPT\footnote{\url{https://chatgpt.com}} and Claude\footnote{\url{https://claude.com/product/overview}}. 
Here, we distill reasoning ability into the smallest model of the Qwen2.5 model family (Qwen2.5-0.5B-Base) via curriculum learning, using information from larger counterparts (1B-32B parameters). Figure \ref{fig:math-family-0.5b} shows that {\levelup} significantly outperforms all other curricula, including the standard practice of training i.i.d. and the C-score baseline. Contrary to this, the {\leveldown} curriculum under-performs {\random} on transitional problems. This indicates that transitional problems are ordered in terms of the \emph{learning difficulty} w.r.t. the model series, which parallels observations in the literature on progressive distillation \citep{mirzadeh2020improved, panigrahi2024progressive}.

\xhdr{Series B: Checkpointing} Frequent checkpointing is another characteristic of training ML models that leads to a model series with a natural progression in strength. To collect our model series, we train the Qwen2.5-1.5B-Base model $M_0$ with 28\% 0-shot test accuracy the GSM8k dataset $\mathcal{D}$, periodically collecting training checkpoints $C_1',C_2',\ldots,C_s'$ that range from 35\% to 59\% validation accuracy. To ensure a meaningful separation between levels, we select a 6-checkpoint subset $\{C_1,C_2,\ldots,C_6\}$ such that there is a notable increase ($\sim 5\%$) in accuracy between each $(C_i,C_{i+1})$ pair on a held-out validation dataset. We then re-train and evaluate copies of Qwen2.5-1.5B-Base on each transitional and baseline curriculum. Figure \ref{fig:math-curr-1.5b} shows that positive results in the distillation setting also hold for this model series. {\levelup} curriculum outperforms {\random} and {\leveldown}, as well as each baseline curriculum. On transitional problems, {\leveldown} underperforms {\random}. We observe a similar trend across multiple experimental settings (Appendix \ref{apx:more_math}). 

\begin{figure}
    \centering
    \begin{minipage}{0.48\textwidth}
        \centering
        \includegraphics[width=\linewidth]{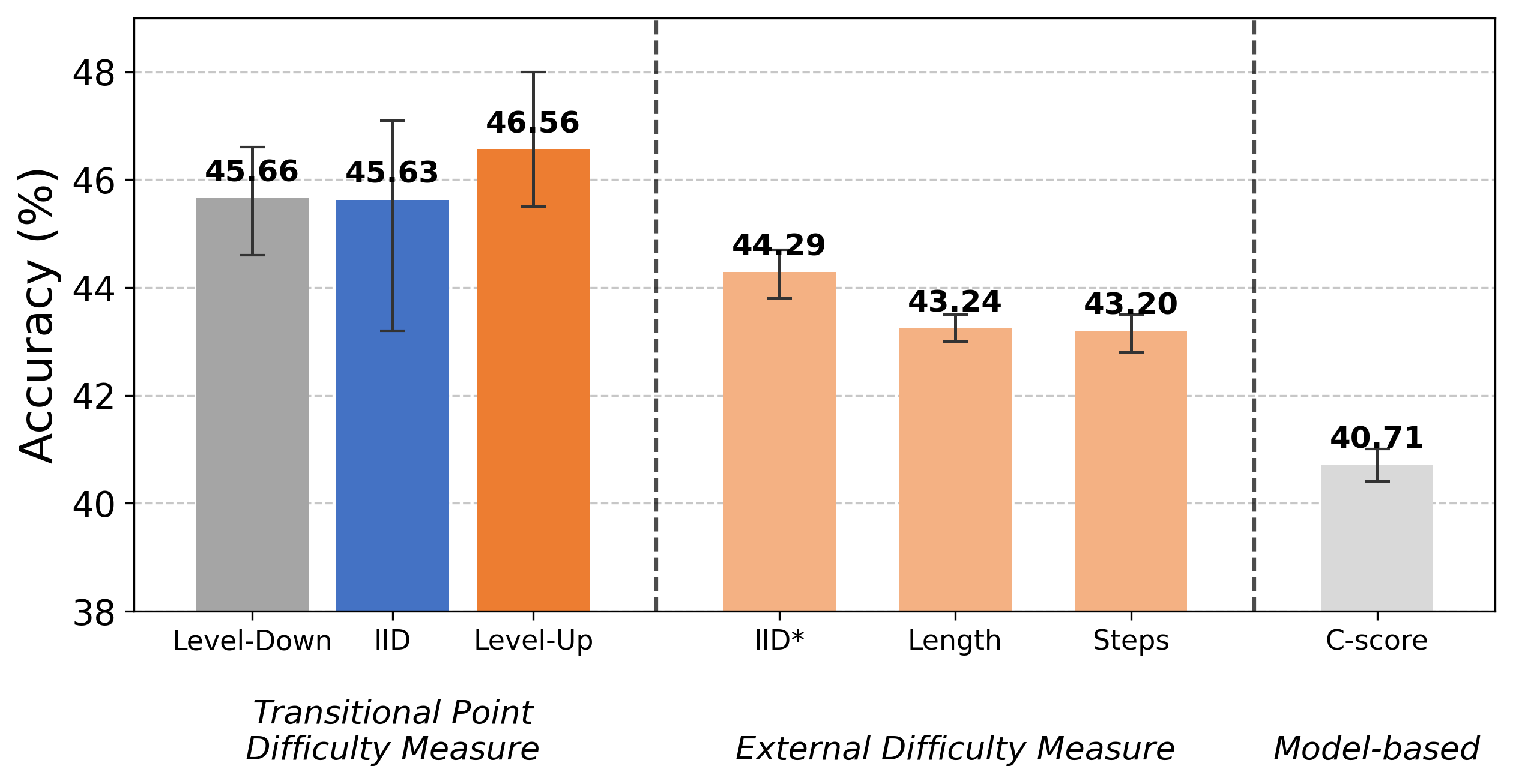}
        \caption{\textbf{Math:} Qwen2.5-1.5B trained on transitional problems obtained from the $k$-shot adaptation model series. {\levelup} still does best, but by a smaller margin than with distillation or full training. Transitional problems appear to be a generally good subset to train on.}\label{fig:math-kshot-1.5}
    \end{minipage}%
    \hfill
    \begin{minipage}{0.43\textwidth}
        \centering
        \includegraphics[width=\linewidth]{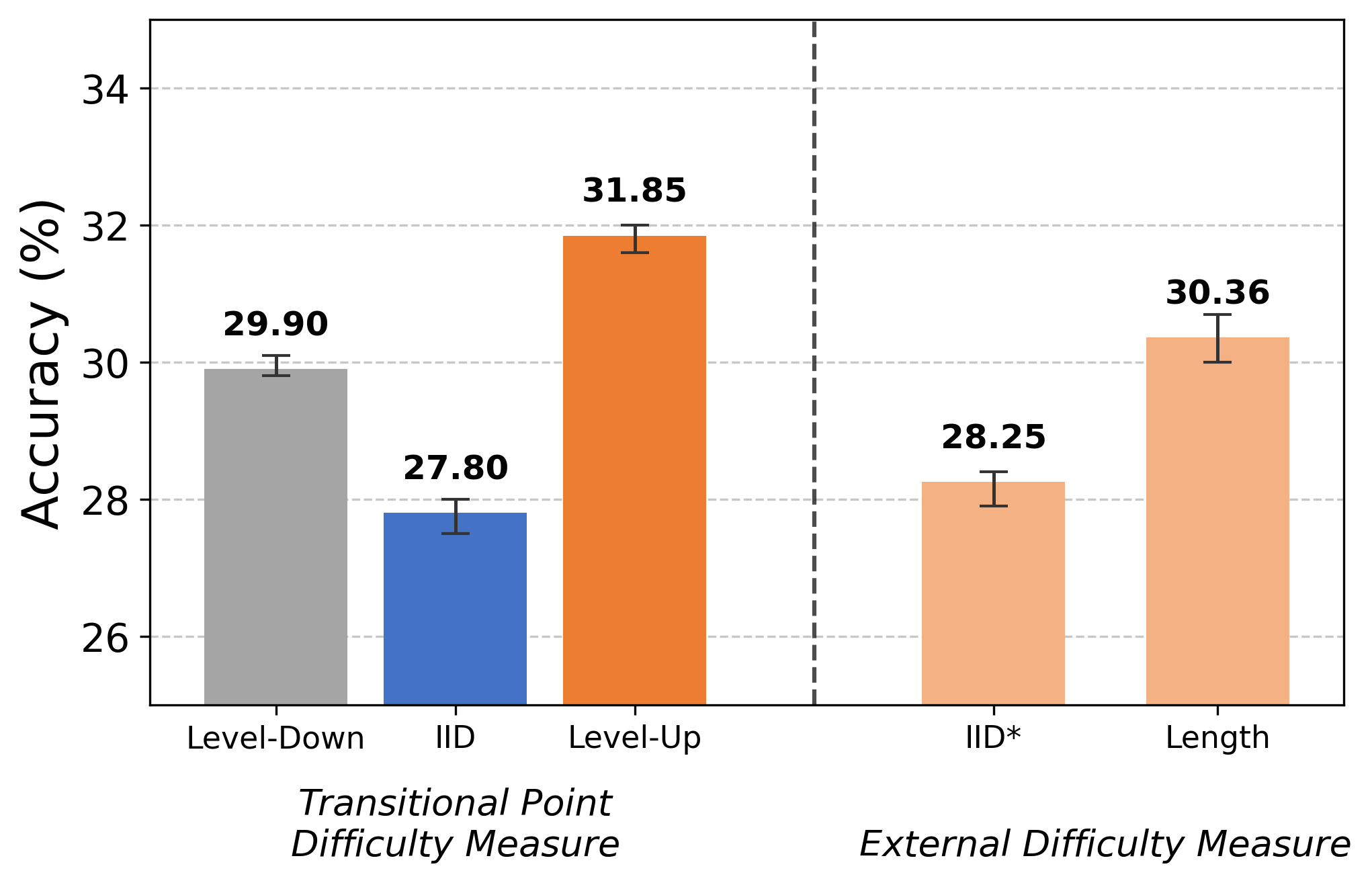}
        \caption{\textbf{Math:} Cross dataset transfer results from GSM8k to Orca-math with Qwen2.5-0.5B trained on {\levelup}, {\random}, and {\leveldown} curricula on \emph{neo-transitional} problems along with Orca baselines. {\levelup} does best.}\label{fig:math-transfer-0.5b-orca}
    \end{minipage}%
% \vspace{-0.5cm}
\end{figure}

\xhdr{Series C: In-context Adaptation} As described in Section \ref{method:model-series}, the strength of an LLM also depends on its task-specific input. Through techniques such as \emph{few-shot in-context adaptation} \cite{brown2020language}, an LLM can leverage context to temporarily improve in strength on a task. Here, we conduct a novel analysis on distilling $k$-shot improvement into a \emph{permanent} increase in task performance with curriculum learning. We collect transitional problems from a 6-model series corresponding to Qwen2.5-1.5B-Base adapted with 0--16 in-context GSM8k examples (with validation accuracies from 35\%--73\% respectively). We then fine-tune the model with each of our curricula, and find again that {\levelup} outperforms every other curriculum (Figure \ref{fig:math-kshot-1.5}). We see a similar result when training the Qwen2.5-0.5B-Base model on these transitional problems, supporting our distillation observations (Appendix \ref{apx:more_math}).

\xhdr{Cross-dataset transfer} Evaluating very large models or collecting checkpoints from task-specific training can be an expensive process. To amortize the training, evaluating, and collecting transitional problems on a large dataset $\mathcal{D'}$, we study a setting where the transitional problems are instead collected on a related but smaller dataset $\mathcal{D}$ and matched level-by-level to identify \emph{neo-transitional} problems on $\mathcal{D'}$. Here, we study the transfer from our GSM8k transitional problems to a formatted version of Orca-200k \citep{mitra2024orcamathunlockingpotentialslms}, a large-scale pretraining dataset of grade-school mathematics word problems. Using a powerful contextual embedding model (Qwen3-Embedding-8B \citep{zhang2025qwen3embeddingadvancingtext}) we embed problems from both datasets into vector space and identify \emph{neo-transitional problems} Orca problems at each level based on their cosine similarity to GSM8k transitional problems at that level. A detailed description of this procedure is presented in Appendix \ref{apx:math_details}. We find that training Qwen2.5-0.5B-Base with the {\levelup} curriculum on \emph{neo-transitional} problems outperforms {\random} and {\leveldown}, as well as two baselines on the Orca training set\footnote{Other baselines do not apply: Orca solutions do not annotate reasoning steps; and we assume no access to model predictions for the C-score.}, showing effective transfer learning (Figure \ref{fig:math-transfer-0.5b-orca}). Note that this result significantly boosts the practicality of our method. By evaluating transitional problems on a small related dataset (e.g., a subset of the full training dataset) and then identifying \emph{neo-transitional} problems on a large training dataset, a curriculum learning method could significantly improve the efficiency of training on the latter while only incurring a minimal additional cost.

\begin{figure}[h]
\centering
\includegraphics[width=\textwidth]{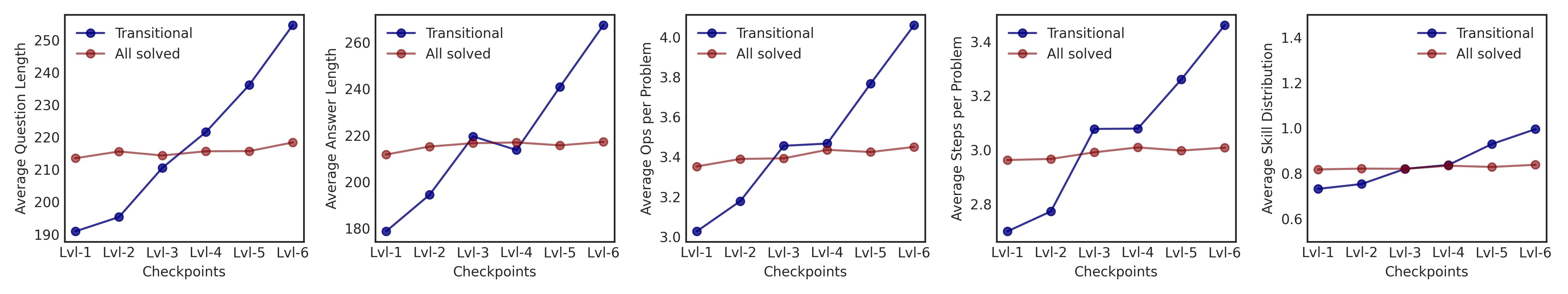}
\caption{\textbf{Math:} Variation in average question length, answer length, and arithmetic operations for transitional problems vs. all solved problems for a 6-model series obtained from the training checkpoints of Qwen2.5-1.5B-Base on GSM8k. }\label{fig:math-interp-compare-solved}
% \vspace{-0.2cm}
\end{figure}

\xhdr{Interpreting Transitional GSM8k Problems} Figure \ref{fig:math-interp-compare-solved} shows that as expected, transitional problems at higher levels of competence are (on average) longer and require more reasoning steps and math operations than transitional problems at lower levels. However, the proportion of arithmetic operations required to solve a problem remains constant across competence levels, despite the division operation being more complex to reason about and implement. Notably, these trends in difficulty are absent across the sets of all problems solved by a model at competence level $i$, indicating that our consistency constraint is crucial in enabling interpretation. We observe the same patterns of alignment across every setting in the math domain, showing that transitional problems naturally engender interpretation (Appendix \ref{apx:more_math}). We present statistics on transitional problem counts in Table \ref{tab:stats}.
\section{Intra-Model Transitional Problems: Chess}
\label{sec:experiments-chess}

In this section, we take a closer look at transitional problems from an \emph{intra-model series}. In addition to a variety of inter-model series, the previous section evaluates an intra-model series constructed by in-context learning. However, in-context learning is typically an artifact of pre-training, and on many tasks, does not lead to consistent or monotonic improvement with increasing adaptation. In contrast, some models are explicitly optimized to modulate their response based on context, providing a more robust and consistent difference in strength across the series. One domain where this is the case is \emph{game-playing}, especially chess, where ML models match the performance of humans at various tiers of competence to emulate competitive sparring partners. To study curricula with transitional problems with such an \emph{intra-model} series, we thus conduct experiments in the \emph{chess} domain. 

\xhdr{Chess as a Model Domain} 
Traditional chess engines such as Stockfish \citep{stockfish} and AlphaZero \citep{silver2017mastering} can vary their `thinking' (search depth) to weaken play from superhuman levels, but this reduction is not well-calibrated and the models cannot then improve by training on chess games. Instead, we use Maia-2~\citep{tang2024maia2}, a state-of-the-art chess foundation model that can mimic human-level chess at various levels of competence.
Maia-2 uses a \emph{skill-aware attention} mechanism to coherently capture the spectrum of human ability, making it well-suited to our study of transitional problems and curriculum learning. Taking as input a chess position and the Elo ratings \citep{elo1978rating} of the active and opposing players, Maia-2 outputs a probability distribution over all legal moves available to the active player. Based on context, the model predicts moves at strengths $M_0\le1100,M_1\le1200,\ldots,M_{10}\le2000$. Our strength function lower bounds this modeling as $s(M_0)=1000,s(M_1)=1100,\ldots,s(M_{10})=1900$. 

\xhdr{Finding Transitional Chess Problems} First, we define the set of \textbf{transitional positions} from regular chess games at transition point $\tau$ as:
\begin{equation}
\mathcal{D}_{\tau}^{{pos}} = \{p\in\mathcal{D}^{pos}:\phi_p(M_i)_{i<\tau}=0\land \phi_p(M_i)_{i\ge\tau}=1\}
\end{equation}
where $\mathcal{D}^{pos}$ is a dataset of positions from human chess games in the \href{https://database.lichess.org/\#evals}{Lichess Database}\footnote{\url{https://lichess.org}} annotated with the best next move according to Stockfish~\citep{stockfish}, and $\phi_p(M_i)$ indicates whether the model at strength $i$ (e.g., 1600 Elo) correctly identifies the strongest move.
While game positions provide diverse training data, \emph{chess puzzles} are particularly high-quality learning material for skill acquisition. Chess puzzles are a subset of regular positions with a unique best move that typically leads to a decisive advantage, specifically designed to isolate and teach critical tactical patterns and strategic concepts. Therefore, we also define the set of \textbf{transitional puzzles} at level $\tau$:
\begin{equation}
\mathcal{D}_{\tau}^{{puz}} = \{p\in\mathcal{D}^{puz}:\phi_p(M_i)_{i<\tau}=0\land \phi_p(M_i)_{i\ge\tau}=1\}
\end{equation}
where $\mathcal{D}^{puz}$ is a set of randomly selected puzzles from \href{https://database.lichess.org/\#puzzles}{Lichess}. Puzzles are a standard way to train human learners on chess tactics, and these transitional puzzle sets enable targeted training on the exact skills and concepts needed by the Maia-2 models to progress from $s_{\tau-1}$ to $s_\tau$.

\begin{figure}
    \centering
    \begin{minipage}{0.48\textwidth}
        \centering
        \includegraphics[width=\linewidth]{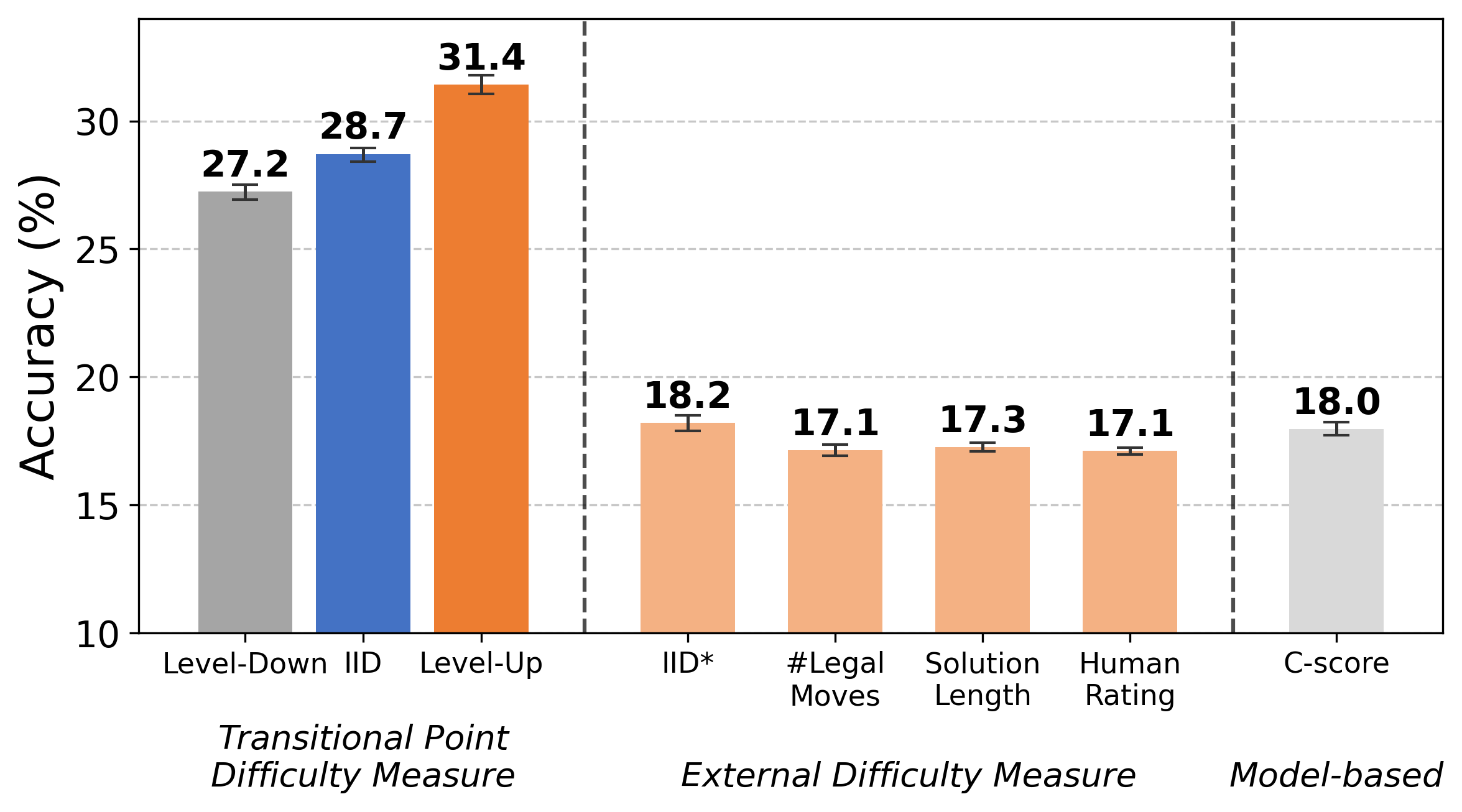}
        \caption{\textbf{Chess:} Maia2-1000 trained on transitional problems from the Maia-2 1100--1900 model series. {\levelup} > {\random} > {\leveldown}, and all of these outperform baselines.}\label{fig:chess-curr-main}
    \end{minipage}%
    \hfill
    \begin{minipage}{0.48\textwidth}
        \centering
        \includegraphics[width=\linewidth]{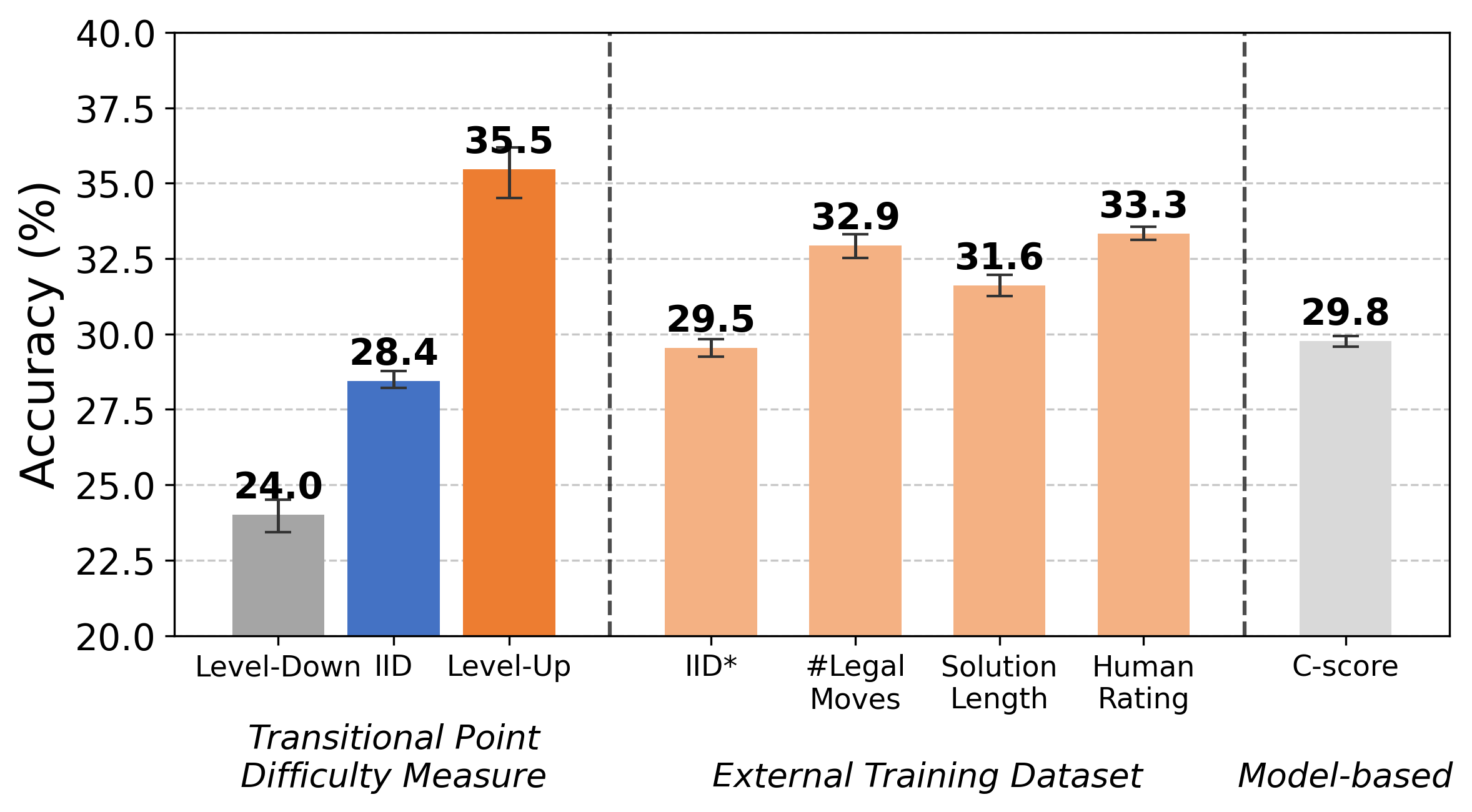}
        \caption{\textbf{Chess:} Transfer from positions to puzzles. Models trained on positions $\mathcal{D}^{pos}$ are evaluated on puzzles $\mathcal{D}^{puz}$. {\levelup} beats strong baselines and other transitional curricula.}\label{fig:chess_transfer}
    \end{minipage}%
% \vspace{-0.3cm}
\end{figure}

\xhdr{1. Learning from transitional chess puzzles} We identify large sets of training and evaluation transitional problems from the Lichess databases using the Maia-2 model series (1100--1900 Elo). We train the weakest setting of the Maia-2 chess foundation model (at 1000 Elo) on the {\levelup}, {\random}, and {\leveldown} curricula. We also train multiple baselines: ordered i.i.d. (random sampling), by the C-score (avg. correct), or by three domain-specific difficulty measures: the \emph{number of legal moves}; \emph{length of the principal variation}; and \emph{player/puzzle ELO rating} in a position. The domain-specific baselines reflect the increase in branching factor, the search depth, and the reasoning difficulty of a chess problem respectively. We first train and evaluate on datasets of chess puzzles, typically used by human learners to improve specific skills. We observe that the {\levelup} transitional curriculum outperforms {\random}, which in turn outperforms {\leveldown} (Figure~\ref{fig:chess-curr-main}). {\levelup} also substantially outperforms all baselines, including i.i.d. training. We find that these patterns hold across a variety of pacing stages and training budgets (Appendix \ref{apx:results}). Additional details of our experiments are presented in Appendix \ref{apx:repro}.

\xhdr{2. Transfer from general positions to chess puzzles} To simulate human improvement on puzzle-solving from general game play, we train models on game-positions ($\mathcal{D}^{pos}$) and evaluate on chess puzzles ($\mathcal{D}^{puz}$)---a structurally similar setting but with differing search characteristics. As shown in Figure~\ref{fig:chess_transfer}, {\levelup} achieves 35.5\% accuracy, outperforming both {\random} and {\leveldown} on transitional data as well as the baselines that use external difficulty measures. This demonstrates that transitional problems generalize beyond specific datasets in similarly-structured domains, indicating robustness to distribution shift.

\begin{figure*}[h]
\centering
\includegraphics[width=0.8\textwidth]{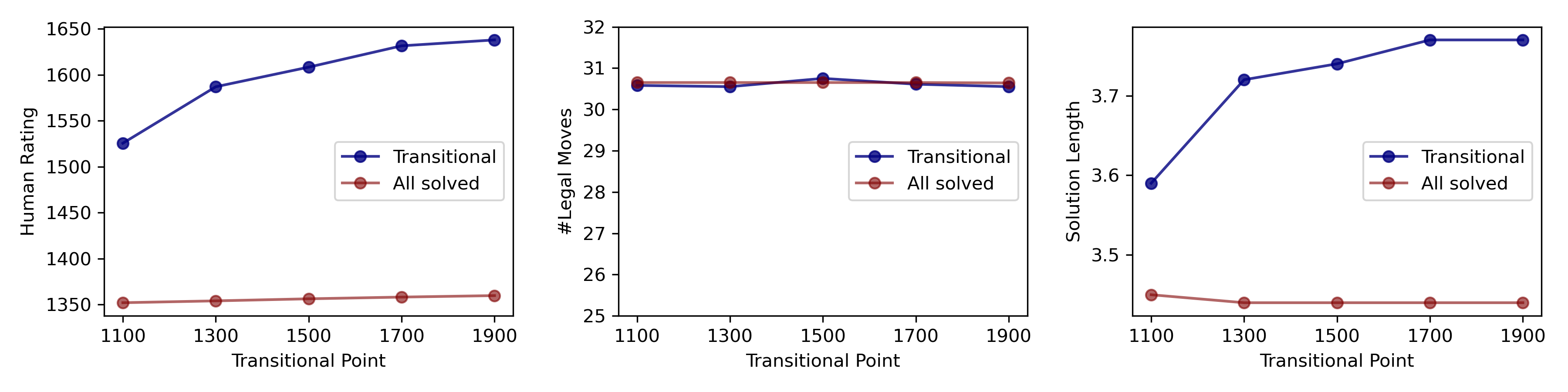}
\caption{
\textbf{Chess}: Average human rating, number of legal moves, and solution length for transitional problems versus all solved problems at each transitional point. For transitional problems, the human rating and solution length correlate with the competence level, while the number of legal moves does not. None of these measures correlate for the full set of puzzles across levels. }\label{fig:chess_correlation}
% \vspace{-0.4cm}
\end{figure*}

\xhdr{Characteristics of Chess Problems} Like in the math setting, we observe that there is virtually no variation in the average difficulty (to humans) of a chess puzzle solved by each level of the Maia-2 model (Figure~\ref{fig:chess_correlation}, in maroon). The Average puzzle rating stays between 1352--1360 Elo, solution length between 3.43--3.45 moves, and the number of legal moves per position at $\sim$30.65. In contrast, the sets of transitional problems (in navy) show a larger increase in the average puzzle rating ($1525\rightarrow 1646$) and solution length ($3.6\rightarrow3.8$), indicating that our consistency constraint leads to better alignment with human difficulty. However, the average number of legal moves remains static even in the latter setting. This may be a characteristic of our dataset (puzzles are usually short and simple setups), but also may indicate that transitional problems cannot be identified simply by composing intuitive difficulty measures. 

\label{sec:atomic_level_up}
\begin{figure*}[h]
\centering
\includegraphics[width=0.9\linewidth]{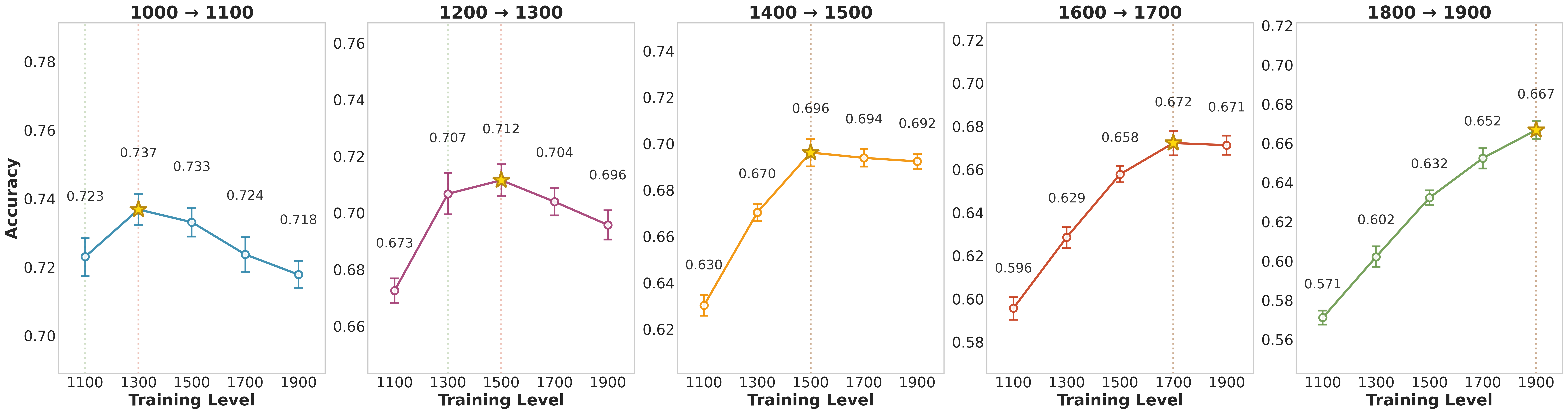}
\caption{\textbf{Chess}: Each subplot depicts results when a chess model at a particular competence level $i$ is trained on puzzles from a single level (along the x-axis), and tested on its ability to solve game-position problems at level $i+1$ (accuracies, y-axis).
The vertical line in each indicates our hypothesized level to achieve the best performance. The star denotes the actual level that achieves the best performance, which largely aligns with our hypothesis. Error bars represent \textit{std} across 10 runs.}\label{fig:chess_level_up}
\end{figure*}

\xhdr{Atomic Leveling Up in Chess} Since we have a very large number of transitional chess puzzles at every competence level (Table~\ref{tab:stats} in the Appendix), we can split them into training and testing subsets and study the effects of curriculum learning in the limiting case. We fine-tune Maia-2 at each level ($M_i=M_{1000}-M_{1800}$) on transitional puzzles at every level, and evaluate which `levels up' the model into solving the most transitional problems at the immediate next level $\mathcal{D}_{i+1}^{puz}$.
As shown in Figure~\ref{fig:chess_level_up}, training on the transitional problems at the immediate next level ($i+1$) almost always produces the maximal improvement in the model at level $i$.
For example, the third subplot shows that $M_{1400}$ performs best on 1500-level testing problems ($\mathcal{D}_{5}^{puz}$) by training with 1500-level problems ($\mathcal{D}_{5}^{puz}$). 
Figure~\ref{fig:chess_level_up_ood} (Appendix \ref{apx:results}) shows similar results for $\mathcal{D}^{pos}$--$\mathcal{D}^{puz}$ transfer learning.

\begin{figure}[htbp]
\centering
\includegraphics[width=0.85\linewidth]{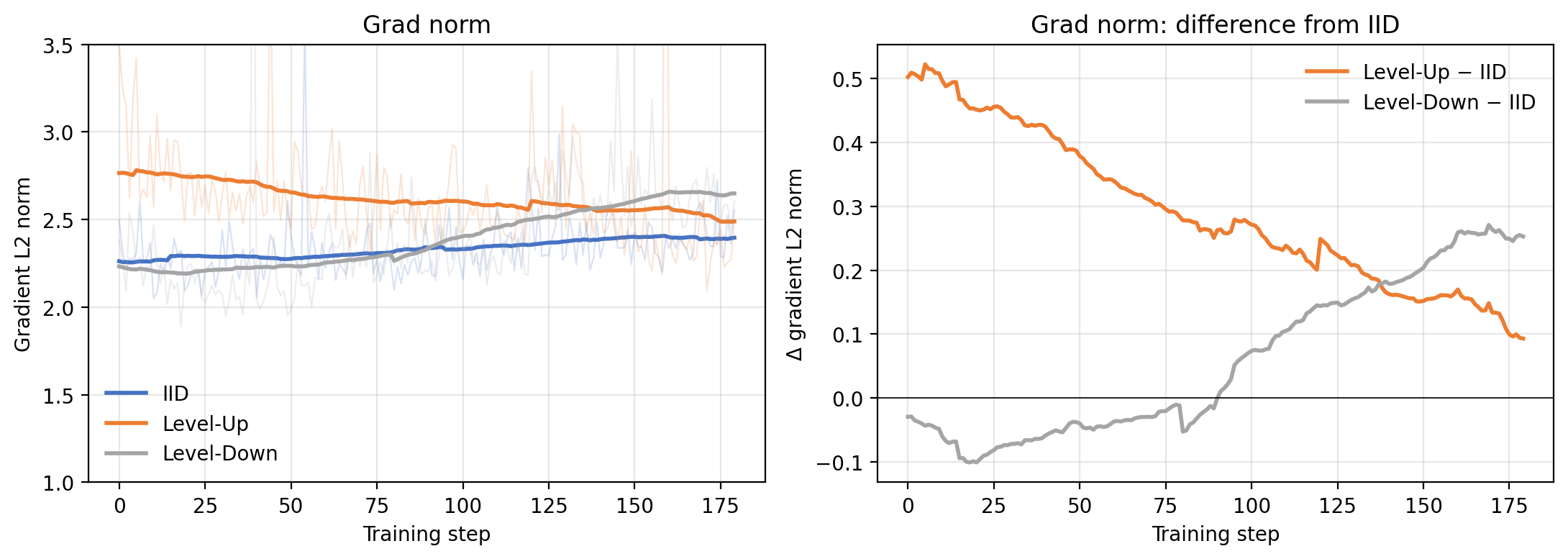}
\caption{Gradient $L^2$ norm of {\levelup}, {\leveldown}, and IID curricula during fine-tuning on chess positions. \textbf{Left:} per-step gradient norm (faint lines) overlaid with a centered rolling mean (solid lines). \textbf{Right:} smoothed difference from the IID baseline.}
\label{fig:grad-norm-curriculum}
\end{figure}

\xhdr{Analyzing the Learning Signal} Inspired by an observation in curriculum learning with infant egocentric videos \cite{sheybani2023curriculum}, we evaluate the magnitude ($L^2$ norm) of Maia-2 gradients during training with {\levelup}, {\leveldown}, and IID curricula. Figure \ref{fig:grad-norm-curriculum} shows that {\levelup} starts with a noticeably larger gradient than IID and decays towards it, while {\leveldown} begins close to IID and grows above it as training progresses. This effect is also observed in the former work, indicating that an effective curriculum exposes the learner to a qualitatively different optimization regime. We note that the learners in both works use a convolutional neural network \cite{NIPS1989_53c3bce6} backbone---we do not observe a similar result in our math setting with LLM learners, likely due to differences in network architecture and output structure.
\section{Discussion}\label{sec:discuss}

In this work, we introduce the notion of \emph{transitional problems} by identifying the training problems that uniquely define the level of competence of an ML model, and are also well-calibrated to models across every level of competence. Through experiments in chess and mathematics, we observe that: (1) \emph{Level-up} curricula that progress through transitional problems of increasing difficulty outperform other curricula on transitional problems and on general training data; and (2) Transitional problems naturally vary along some human measures of problem difficulty, despite not being explicitly optimized to do so. Thus, transitional problems are a \emph{model-specific easy-to-hard} ordering of a subset of training problems based on learning difficulty.

{\bf Limitations.} Our positive preliminary results show that {\levelup} curriculum learning on transitional problems is potentially a strong and widely applicable training method. However, we only evaluate a constant pacing strategy for each curriculum, and do not mix data between blocks of training. In future work, we aim to explore the interaction of these factors with transitional curricula, and explore large-scale and dynamic curriculum learning with transitional problems in depth.

\section*{Acknowledgement}
This work is supported by the funds provided by the National Science Foundation and by DoD OUSD (R\&E) under Cooperative Agreement PHY-2229929 (The NSF AI Institute for Artificial and Natural Intelligence).
We also acknowledge the John
Templeton Foundation (Grant 63578) and ONR Grant N00014-23-1-2436 for their generous support.  
We thank the members of Zemel Group whose thoughtful feedback helped improve this work.

\section*{Impact Statement}
This paper presents work whose goal is to advance the field of Machine Learning. Our work draws inspiration from theories of human learning, specifically the Zone of Proximal Development, to develop more principled approaches to curriculum learning. While we focus on mathematical reasoning and chess, the broader application of learner-aware difficulty measures to foundation model training could influence how future models are developed. There are many potential societal consequences of our work, none which we feel must be specifically highlighted here.

% \clearpage
\bibliography{references}
\bibliographystyle{abbrv}

%%%%%%%%%%%%%%%%%%%%%%%%%%%%%%%%%%%%%%%%%%%%%%%%%%%%%%%%%%%%

\clearpage
\appendix

\section{Additional Related Work}\label{apx:more-related}

\paragraph{Structured Training} The two major approaches to structured training are \emph{active learning} \citep{cohn1994improving} and \emph{curriculum learning}. Active learning optimizes for the \emph{short-term} goal of finding the best next problem(s) to train a learning model on. Strategies for active learning generally rely on `teacher' models to select batches that minimize the uncertainty over the training distribution, via data clustering \citep{citovsky2021batch}, influence functions \citep{Liu_2021_ICCV}, and coresets \citep{sener2017active}. These methods tend to add considerable overhead and require strong estimates of learner uncertainty, and are thus not popular for training massive-scale foundation models. Contrary to this approach, curriculum learning optimizes for the \emph{long-term} of maximizing the performance at the end of a training phase instead of per-step uncertainty. These approaches can be combined to develop curricula with dynamic batch selection using teacher-student models \citep{matiisen2019teacher} or multi-armed bandits \citep{graves2017automated}, but this adds considerable overhead.

\paragraph{Curriculum Learning in the Literature}

Though ERM remains the de facto recommendation in ML, we observe that training with various forms of non-uniform data ordering is standard in modern massive-scale ML pipelines. LLMs are frequently pre-trained on sequences that increase in length and context windows that grow in size over training \citep{liu2024deepseek}. Similarly, curricula are also useful in learning other long sequences of data with autoregressive models, at scales ranging from simple parity tasks for RNNs \citep{khajehabdollahi2023emergent} to peptide sequencing \citep{zhang2025curriculum}, graph learning \citep{li2025self}, and diffusion models \citep{kim2024denoising}.
Reinforcement learning methods often use curricula to learn increasingly long action sequences \citep{narvekar2020curriculum, patel2402curriculum, li2025causally, zhao2022learning} and for faster convergence \citep{tao2024reverse}. 
The training of many autoregressive foundation models has a consistent, curriculum-like structure, with large-scale pre-training phase on  unstructured data followed by a structured post-training phase \cite{brown2020language}. Recent training schemes include a `mid-training' phase \citep{wake2024yi, olmo20242} that focuses on enhancing capabilities with higher-quality data than in pre-training. The highly structured post-training phase has shown wins for curricula in training for instruction-following \citep{ge2025dynamic} and complex reasoning \cite{zeng2025glm, polu2022formal}. In contrast, efforts to explicitly incorporate curriculum learning into pre-training have been largely unsuccessful, even with data inspired by human development. The BabyLM challenge \citep{oba2023babylm} saw a myriad of curriculum-based approaches fail to beat the baseline for pre-training a language model on 10-100 million tokens of data. These included domain-specific \citep{martinez2023climb, edman2023too, oba2023babylm}, model-dependent \citep{opper2023effect}, and student-teacher \citep{chobey2023can, zhang2023baby} setups, but notably did not evaluate curricula that dynamically assess example difficulty to determine the next inputs during training. Other negative results for curriculum learning stem from settings where models are overparameterized \citep{mannelli2024tilting} or can be trained to convergence with relatively noise-free datasets \citep{wu2020curricula}.

\section{Additional Chess Results}\label{apx:results}

\xhdr{Evaluating Maia-2 models, in detail}

In Section \ref{sec:method}, we provide an overview of our process for collecting transitional problems with Maia-2. Since this model is uniquely suited to the chess domain, we outline our exact methodology for this process below. 

We set both the active and opponent strengths to $s_i$ to ensure a consistent skill representation. 
For a given chess position $b$ conditioned on $s_i$, the Maia-2 policy head output
$f_\theta(b, s_i) = \pi(\mathbf{a}|b, s_i)$,
estimates the probability distribution over legal moves (i.e., actions) $\mathbf{a}$.
Each $M_i$ can be fine-tuned with the following loss function:
\begin{equation}
\mathcal{L}(\theta) = -\mathbb{E}_{b \sim \mathcal{D}} [\log \pi(a^* | b, s_i)]
\end{equation}
where $\mathbb{E}_{b \sim \mathcal{D}}[\cdot]$ denotes the expectation over chess position $b$ sampled from the data distribution $\mathcal{D}$, where $a^*$ is the target move under $b$, which we set to be the highest-ranked move according to Stockfish. This serves as a proxy for the best move, and in positions where multiple moves are equally strong, this choice is arbitrary.
We determine move correctness as:
\begin{equation}
\phi_p(M_i) = \begin{cases}
1 & \text{if } \operatorname*{argmax}_{\mathbf{a}} \pi(\mathbf{a}|b, s_i) = a^* \\
0 & \text{otherwise}
\end{cases}
\end{equation}
This allows us to map each problem $p = \{b, a^*\}$ to its transition point $\tau_p \in \{1, 2, \ldots, 10\}$ in the model series, where positions solvable or unsolvable by all models are excluded.

\begin{figure}[hb]
    \centering
    \includegraphics[width=0.6\linewidth]{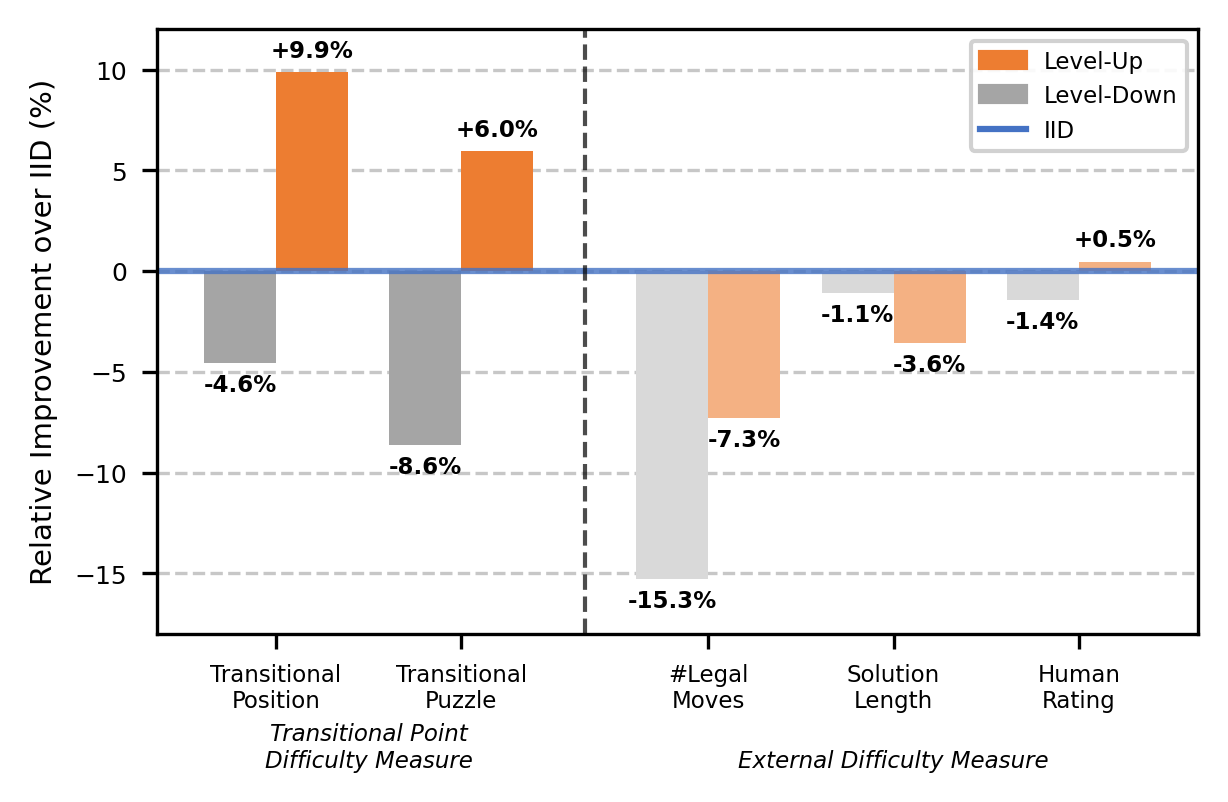}
    \caption{Relative improvement over IID on chess, evaluated on test sets matched to each difficulty measure. Only transitional point-based measures (left) benefit from Level-Up curriculum ordering, while external difficulty measures (right) show marginal improvement or degradation.}
    \label{fig:chess_main}
\end{figure}

\xhdr{Transitional Problems beget Curriculum learning} As shown in Figure~\ref{fig:chess_main}, only training on transitional problems yields positive results for easy-to-hard (i.e., level-up) curricula: Transitional Position achieves +9.9\% and Transitional Puzzle achieves +6.0\% improvement over IID. In contrast, external difficulty measures fail to benefit from a curriculum/anti-curriculum, with \#Legal Moves showing substantial degradation (-7.3\%) and Solution Length and Human Rating remaining roughly on par with IID. This indicates that transitional problems capture a notion of difficulty that is particularly amenable to curriculum learning, whereas external heuristics do not induce a useful easy-to-hard structure for the learner.

\begin{figure}[htbp]
    \centering
    \begin{minipage}{0.48\linewidth}
        \centering
        \includegraphics[width=\linewidth]{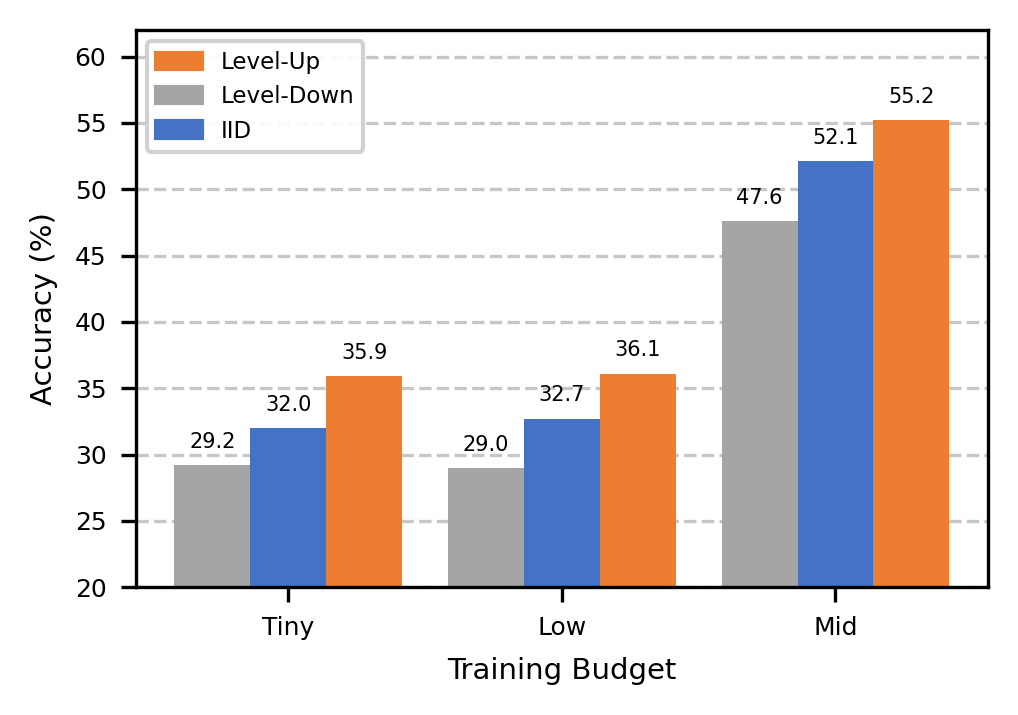}
        \caption{Performance comparison of curriculum learning strategies for chess models across different computational budgets. IID, Level-Up, and Level-Down denote the i.i.d baseline, easy-to-hard curriculum, and hard-to-easy curriculum, respectively, with the Tiny, Low, or Mid training budgets defined in Table~\ref{tab:hyperparameters_chess}.}
        \label{fig:chess_budget}
    \end{minipage}
    \hfill
    \begin{minipage}{0.48\linewidth}
        \centering
        \includegraphics[width=\linewidth]{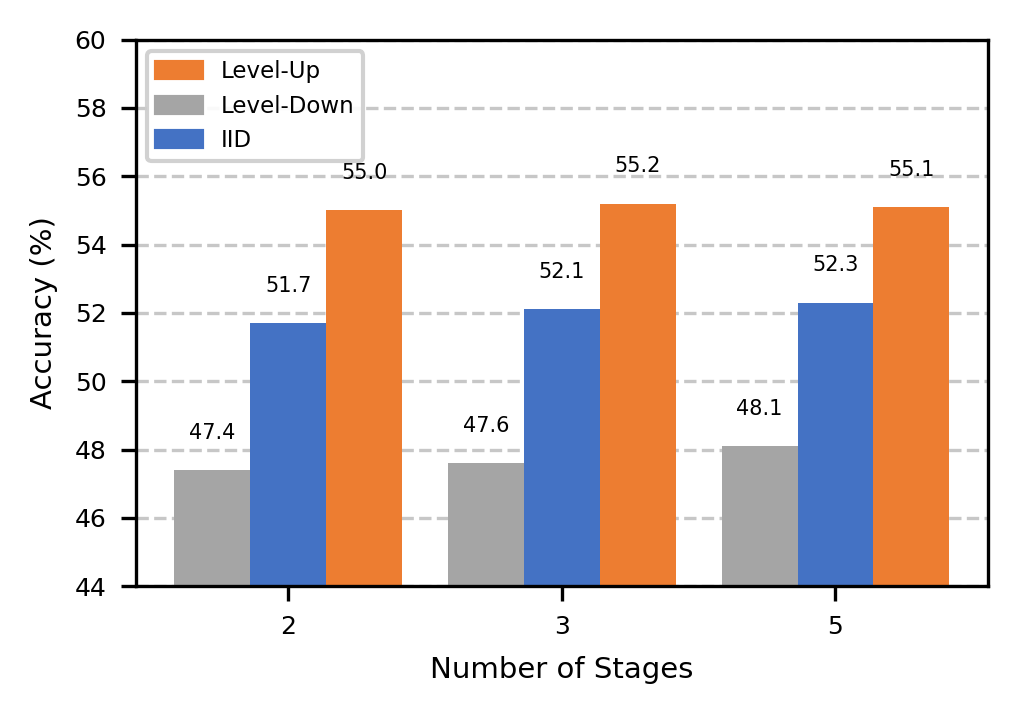}
        \caption{Performance comparison of curriculum learning strategies for chess models across the number of curriculum steps. IID, Level-Up, and Level-Down. denote the i.i.d baseline, ascending curriculum from easy to hard, and descending curriculum from hard to easy, respectively, including 2, 3, or 5 curriculum steps.}
        \label{fig:chess_stage}
    \end{minipage}
\end{figure}

\xhdr{Effect of Training Budgets}
As shown in Figure~\ref{fig:chess_budget}, the advantage of Level-Up curricula holds consistently across training budgets. Level-Up outperforms both IID and Level-Down at all scales (Tiny, Low, Mid), with the relative ordering Level-Up \textgreater{} IID \textgreater{} Level-Down preserved throughout. The absolute gap between Level-Up and IID remains stable ($\sim$3--4\%), indicating that the benefits of transitional point-based curricula are robust across different computational budgets.

\xhdr{Effect of Curriculum Steps}
As shown in Figure~\ref{fig:chess_stage}, we did not find notable patterns in the effect of curriculum steps in chess. 
We would like to clarify that our goal to include multiple settings in terms of steps is to show the \emph{consistency} of the advantage of level-up curricula over baselines, instead of finding the best hyperparameter setting. 

\begin{figure}[h]
\centering
\includegraphics[width=\linewidth]{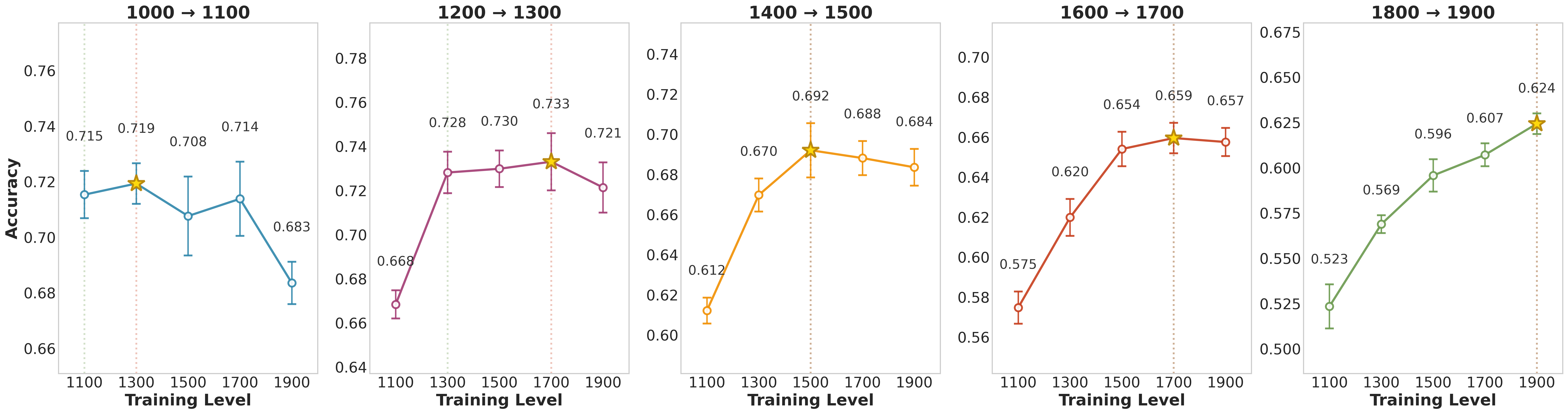}
\caption{Each subplot depicts results when a \textbf{chess} model at a particular competence level $i$ is trained on game-positions from a single level (along the x-axis), and tested on its ability to solve puzzles at level $i+1$ (accuracies, y-axis).
The vertical line in each indicates our hypothesized level to achieve the best performance. The star denotes the actual level that achieves the best performance, which largely aligns with our hypothesis. Error bars represent \textit{std} across 10 runs.}\label{fig:chess_level_up_ood}
\end{figure}

\xhdr{Transfer Learning in Atomic Level-Up}
We have shown the results with $\mathcal{D}^{puz}$-$\mathcal{D}^{puz}$ pair ($\mathcal{D}^{puz}$ for training, $\mathcal{D}^{puz}$ for testing) in the main paper, while we also observe consistent patterns with the $\mathcal{D}^{pos}$ - $\mathcal{D}^{puz}$ pair, which is under such a \emph{transfer learning} setting.
We fine-tune models $\{M_0, M_2, M_4, M_6, M_8\}$ on their `level-up' transitional puzzles $\mathcal{D}_{\tau}^{pos}, \tau \in \{1, 3, 5, 7, 9\}$ and test on the equivalent transitional puzzles $\mathcal{D}_{i+1}^{puz}$ for model $M_i$. 
As the results show in Figure~\ref{fig:chess_level_up_ood}, training on transitional problems one level up from the model's competence lead to the largest improvement in most settings.

% \begin{figure}[htbp]
% \centering
% \includegraphics[width=\linewidth]{figures/grad_norm_curriculum.png}
% \caption{Gradient L2 norm during fine-tuning on chess positions for IID, Level-Up, and Level-Down. \textbf{Left:} per-step gradient norm (faint lines) overlaid with a centered rolling mean (solid lines). \textbf{Right:} smoothed difference from the IID baseline. {\levelup} starts with a noticeably larger gradient than IID and decays towards it, while {\leveldown} begins close to IID and grows above it as training progresses, indicating that each exposes the model to a qualitatively different optimization regime. A similar effect is observed in curriculum learning with neural networks on infant videos \cite{sheybani2023curriculum}.}
% \label{fig:grad-norm-curriculum}
% \end{figure}

\section{Additional Math Results}\label{apx:more_math}

In Section \ref{sec:experiments-math}, we examine the effect of training the Qwen2.5-1.5B-Base model on GSM8k transitional problems in an \emph{in-distribution} setting. We also evaluate a progressive distillation setting, where the Qwen2.5-0.5B model is trained on larger models from its model family. We find that in these settings, the `level-up' (easy-to-hard via increasing transitional problem level) curriculum outperforms the random and level-down curricula, and also outperforms curricula on the original training distribution for the same amount of compute. In this section, we evaluate curriculum learning with transitional problems in additional settings, and find that our results generalize beyond the specific model and collection method combinations presented in the main text.

\subsection{Transitional Problems from the Trained Qwen2.5-0.5B Series}

\begin{figure}
    \centering
    \begin{minipage}{0.48\textwidth}
        \centering
        \includegraphics[width=\linewidth]{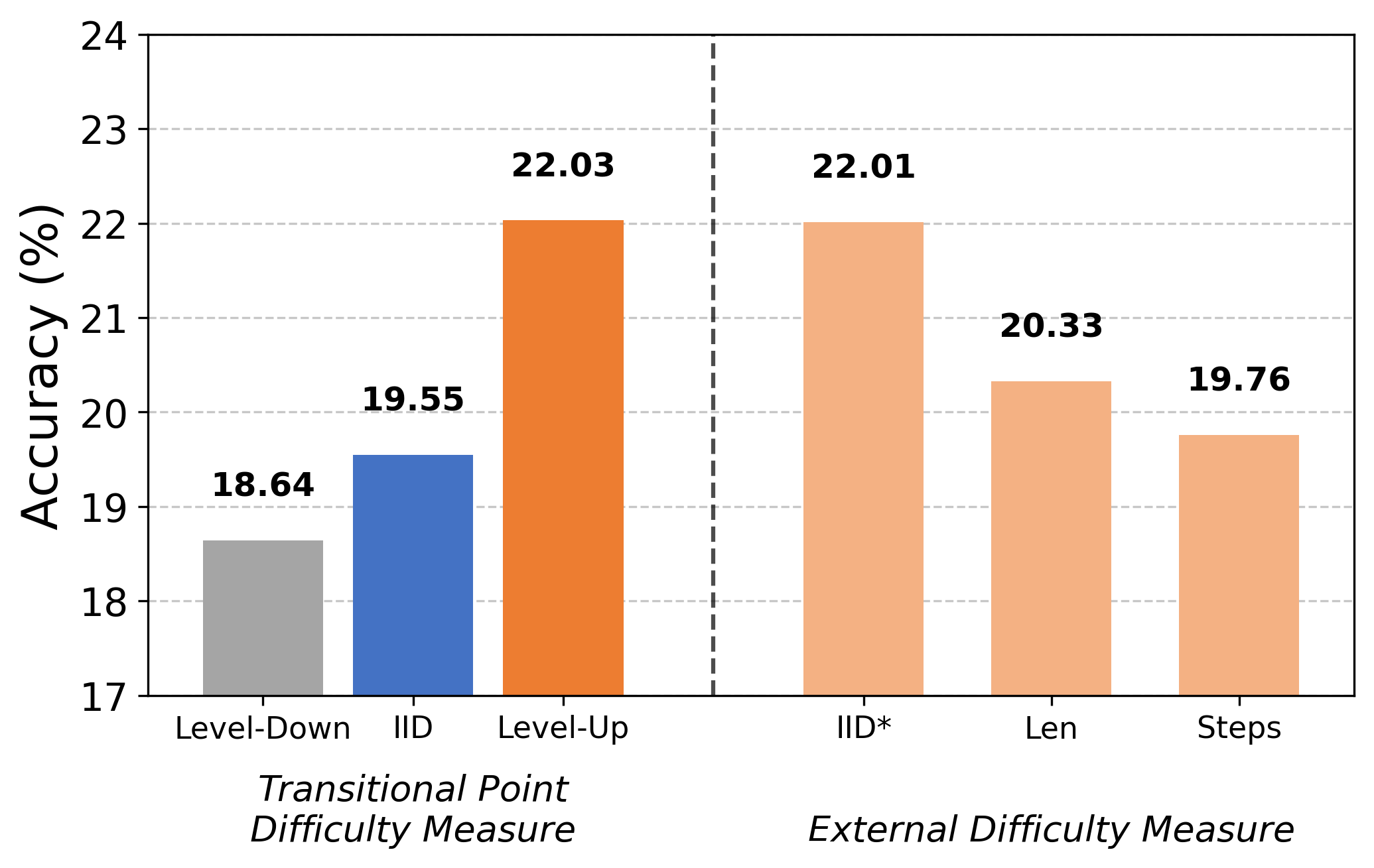}
        \caption{Qwen2.5-0.5B-Base trained on transitional problems from the few-shot adapted Qwen2.5-1.5B model series. The level-up curriculum outperforms every other curriculum, but only shows a slim win over training i.i.d. on the full dataset.}
        \label{fig:math-curr-0.5b}
    \end{minipage}%
    \hfill
    \begin{minipage}{0.48\textwidth}
        \centering
        \includegraphics[width=\linewidth]{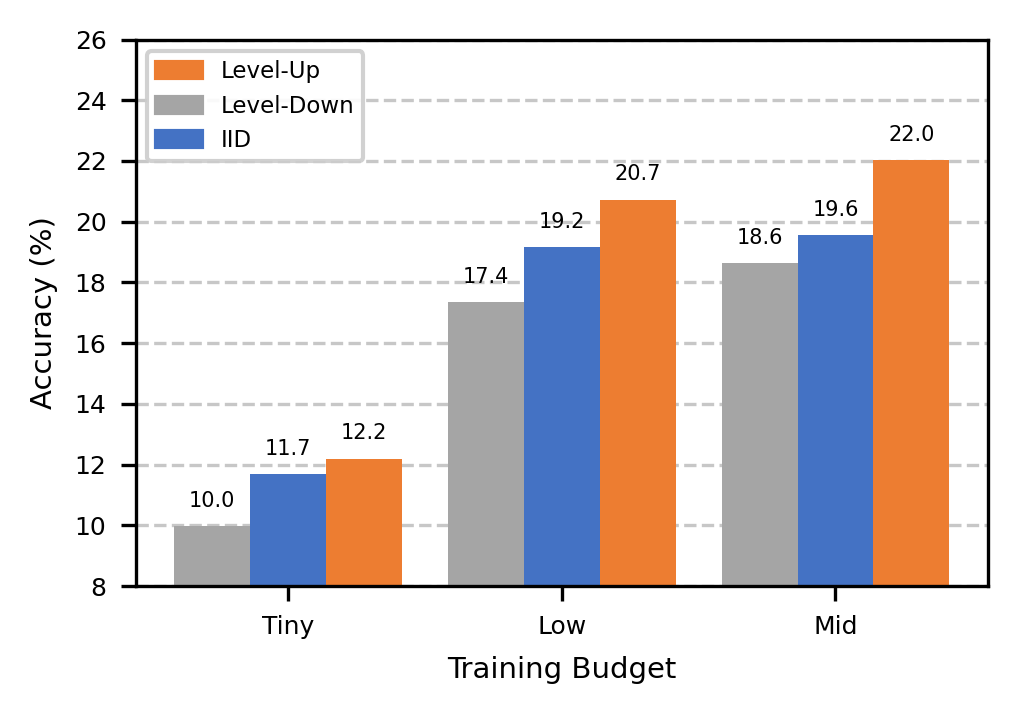}
        \caption{As the number of training steps increases, the level-up curriculum maintains an edge over other curricula on transitional problems, up until the saturation of test set performance.}\label{fig:math-curr-0.5b-steps}
    \end{minipage}%
\end{figure}

Section \ref{expt:math-curr} showed the effect of training Qwen2.5-1.5B-Base on a model series that is a collection of selected checkpoints from the training and evaluation of a separate copy of Qwen2.5-1.5B-Base. Figure \ref{fig:math-curr-0.5b} shows a similar result for this experiment conducted with the Qwen2.5-0.5B-Base model. In this case, the level-up curriculum outperforms other curricula, but only matches the performance of i.i.d. sampling from the full training distribution. Nonetheless, this reinforces our result that transitional problems represent an easy-to-hard ordering of training problems based on learning difficulty, as the level-down curriculum performs worse than a random mixture of transitional problems, and the level-up curriculum performs significantly better than random. Figure \ref{fig:math-curr-0.5b-steps} shows that the level-up curriculum beats the random and level-down curricula across training budgets.

\begin{figure}[h]
\centering
\includegraphics[width=\linewidth]{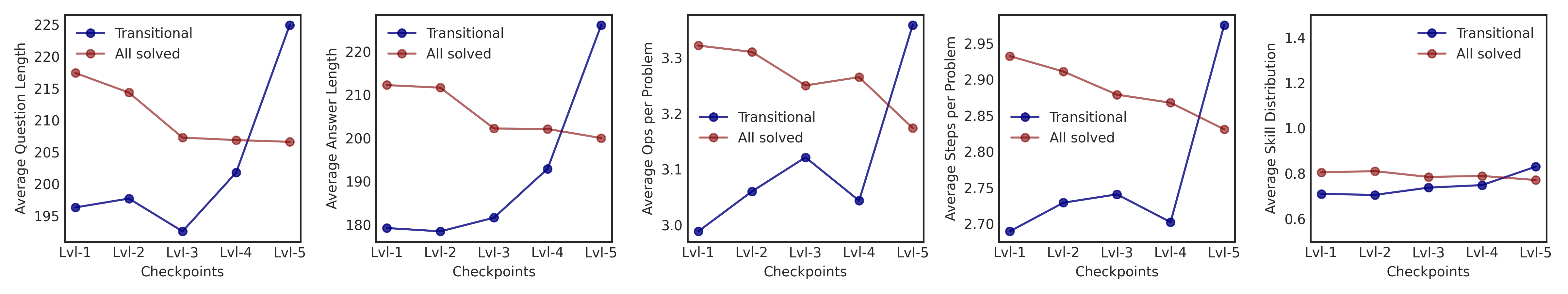}
\caption{Variation in average question length, answer length, and arithmetic operations for transitional problems vs. all solved problems for a 5-model series obtained from the training checkpoints of \textbf{Qwen2.5-0.5B} on GSM8k.}\label{fig:math-interp-0.5b_5}
\end{figure}

\begin{figure}[h]
\centering
\includegraphics[width=\linewidth]{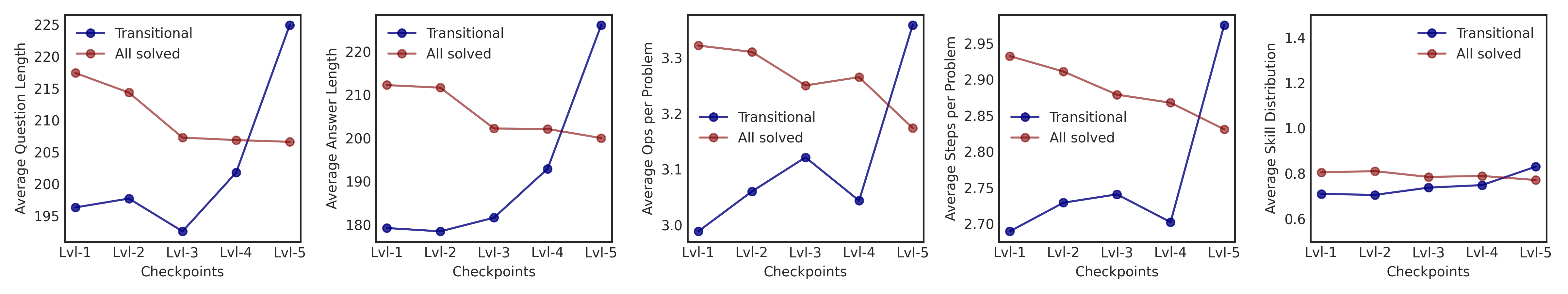}
\caption{Variation in human measures of difficulty for problems obtained from a 4-model series obtained from the training checkpoints of \textbf{Qwen2.5-1.5B} on GSM8k. Like for the 6-model series, the transitional problems vary much more sharply across human-interpretable features than the set of all solved problems.}\label{fig:math-interp-1.5b_4}
\end{figure}

\xhdr{Interpretability} As with the the Qwen2.5-1.5B checkpoint series, we observe that the transitional problems derived from the 5-model series on Qwen2.5-0.5B varies by level along human intuition in the length and the number of steps in the answer, but not in the type of skills required to solve a problem on average (Figure \ref{fig:math-interp-0.5b_5}). This trend is weaker than in the series from 1.5 billion parameter model, which may explain the relatively weaker performance relative to i.i.d. training on the full data distribution that is observed in Figure \ref{fig:math-curr-0.5b-steps}.

Additionally, we show that this variation in difficulty is not unique to any given set of checkpoints (i.e., one model series) from training a base model. Figure \ref{fig:math-interp-1.5b_4} shows that a 4-model series (with ~10\% improvement in accuracy between levels) shows the same trends as a 6-model series (where the improvement drops to ~5\%, Figure \ref{fig:math-interp-compare-solved}), indicating that our curriculum results are likely to hold for any definition of a model series with a reasonable difference in performance between levels.

\subsection{Additional Results on Progressive Distillation}

\begin{figure}
    \centering
    \begin{minipage}{0.48\textwidth}
        \centering
        \includegraphics[width=\linewidth]{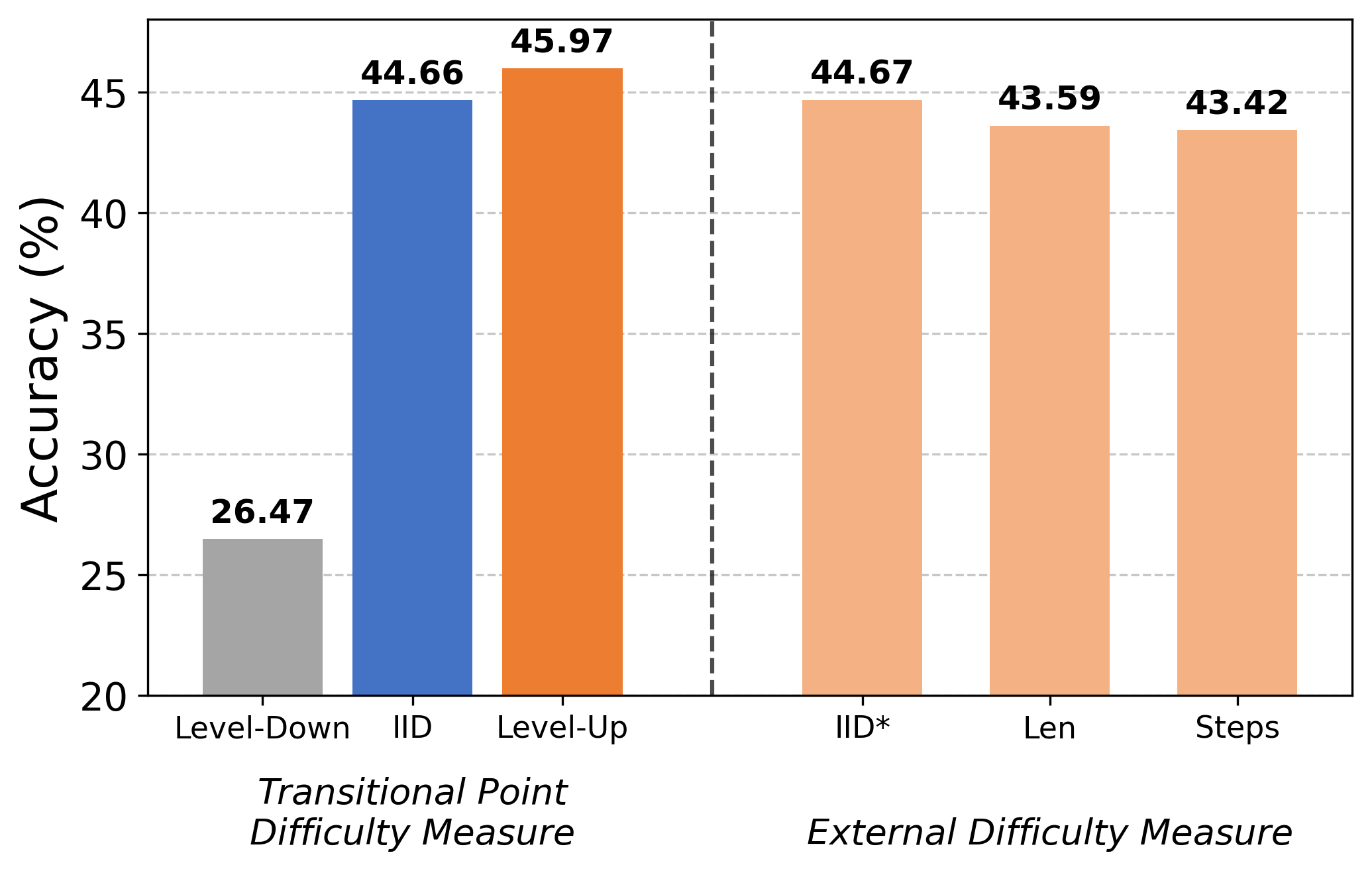}
        \caption{Qwen2.5-1.5B trained on transitional problems from a models series consisting of the 7, 14, and 32 billion parameter Qwen2.5 models.}
        \label{fig:math-curr-family-1.5b}
    \end{minipage}%
    \hfill
    \begin{minipage}{0.48\textwidth}
        \centering
        \includegraphics[width=\linewidth]{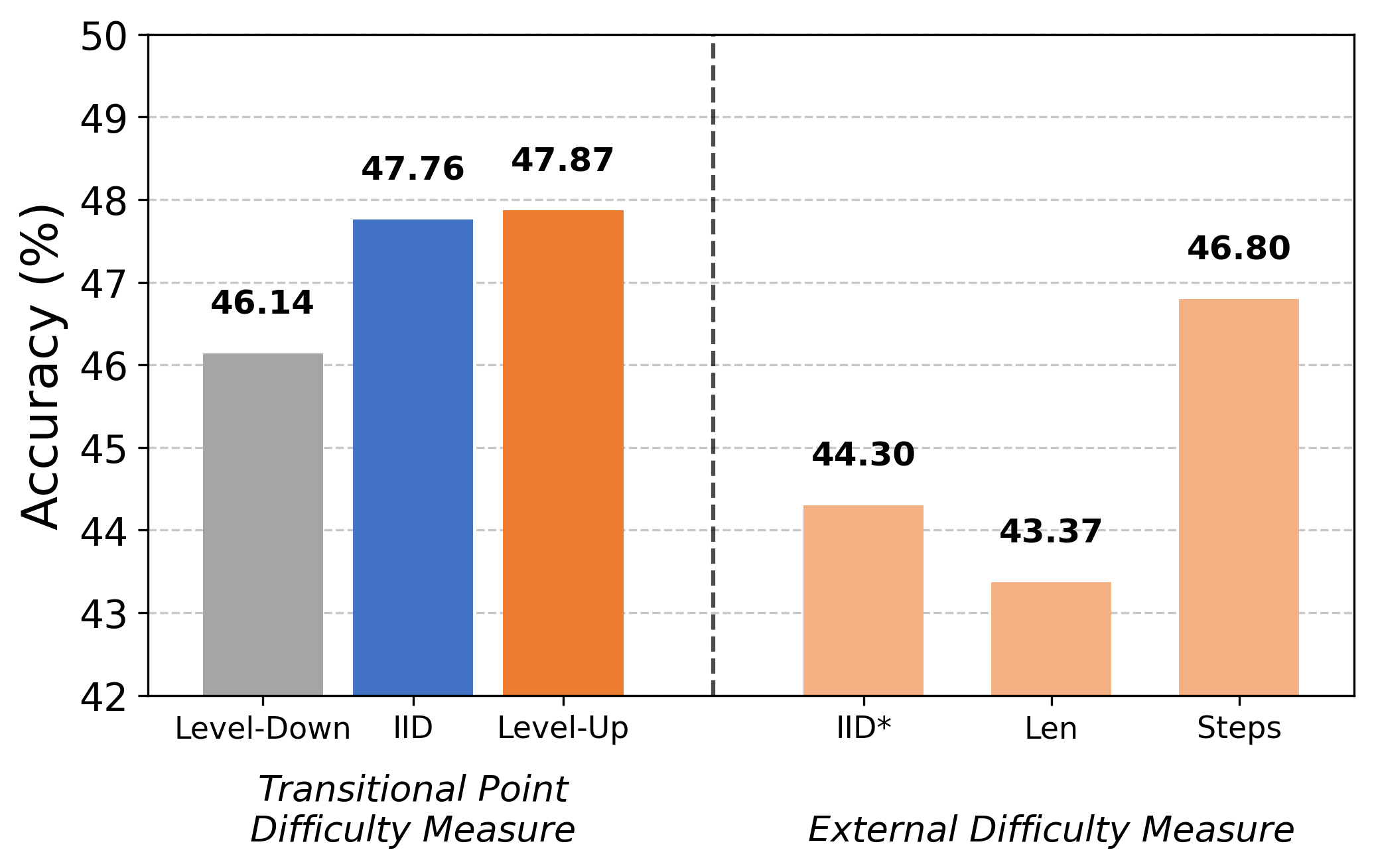}
        \caption{As the number of training steps increases, the lack of additional levels (and thus, transitional problems) reduces the advantage of the level-up curriculum maintains an edge over other curricula.}\label{fig:math-curr-family-1.5b-big}
    \end{minipage}%
\end{figure}

In Section \ref{subsec:expt-math-family}, we show that transitional problems can be used as an instrument of progressive distillation---transferring knowledge from more competent (here, larger) models to less competent (smaller) models by training the Qwen2.5-0.5B base model on transitional problems from a model series corresponding to increasing model sizes (1.5B, 7B, 14B, 32B) in the Qwen2.5 family of models. Figure \ref{fig:math-curr-family-1.5b} shows the effect of training Qwen2.5-1.5B on the subsequently stronger models, with the level-up curriculum corresponding to the transitional problems from the $7\rightarrow 14\rightarrow 32$ billion levels, respectively. As in all other settings, the level-up curriculum outperforms the random and level-down curricula as well as the training-set baselines. However, the margin of improvement is smaller than with Qwen2.5-0.5B, particularly with a few more training steps (Figure \ref{fig:math-curr-family-1.5b-big}), likely due to the reduced number of levels (3 as opposed to 4) due to the unavailability of a larger model in the same family.

\begin{figure}[h]
\centering
\includegraphics[width=\linewidth]{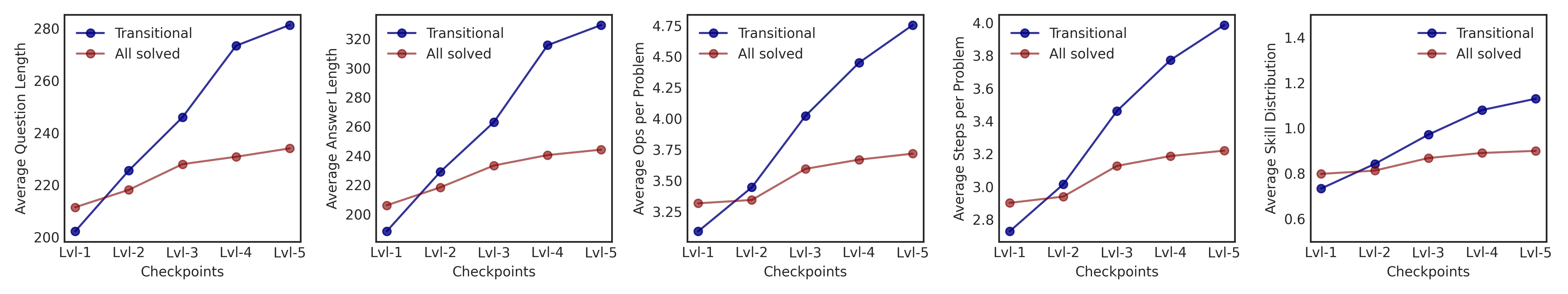}
\caption{Variation in human measures of difficulty for a 5-model series obtained from the Qwen2.5 model family. Due to the large improvement in performance between levels, the set of all solved problems also shows an increase in difficulty. However, the transitional problems show a much sharper change between levels, indicating better differentiation in learning difficulty.}\label{fig:math-interp-family_5}
\end{figure}

\xhdr{Interpretability} The performance of each model in the Qwen2.5 model family (i.e., model series) is significantly higher ($>10\%$) than its smaller predecessor. Thus, we find that even the sets of all solved problems by a model at level $i$ vary mildly along human-interpretable criteria. However, as seen in Figure \ref{fig:math-interp-family_5}, the transitional problems still show a much stronger increase in difficulty (by, e.g., question or answer length) than the set of all solved problems, showing that the consistency constraint provides a useful separation even when the gap between levels is large.

\subsection{An Alternate Formulation: the $k$-shot Model Series}

\begin{figure}
    \centering
    \begin{minipage}{0.48\textwidth}
        \centering
        \includegraphics[width=\linewidth]{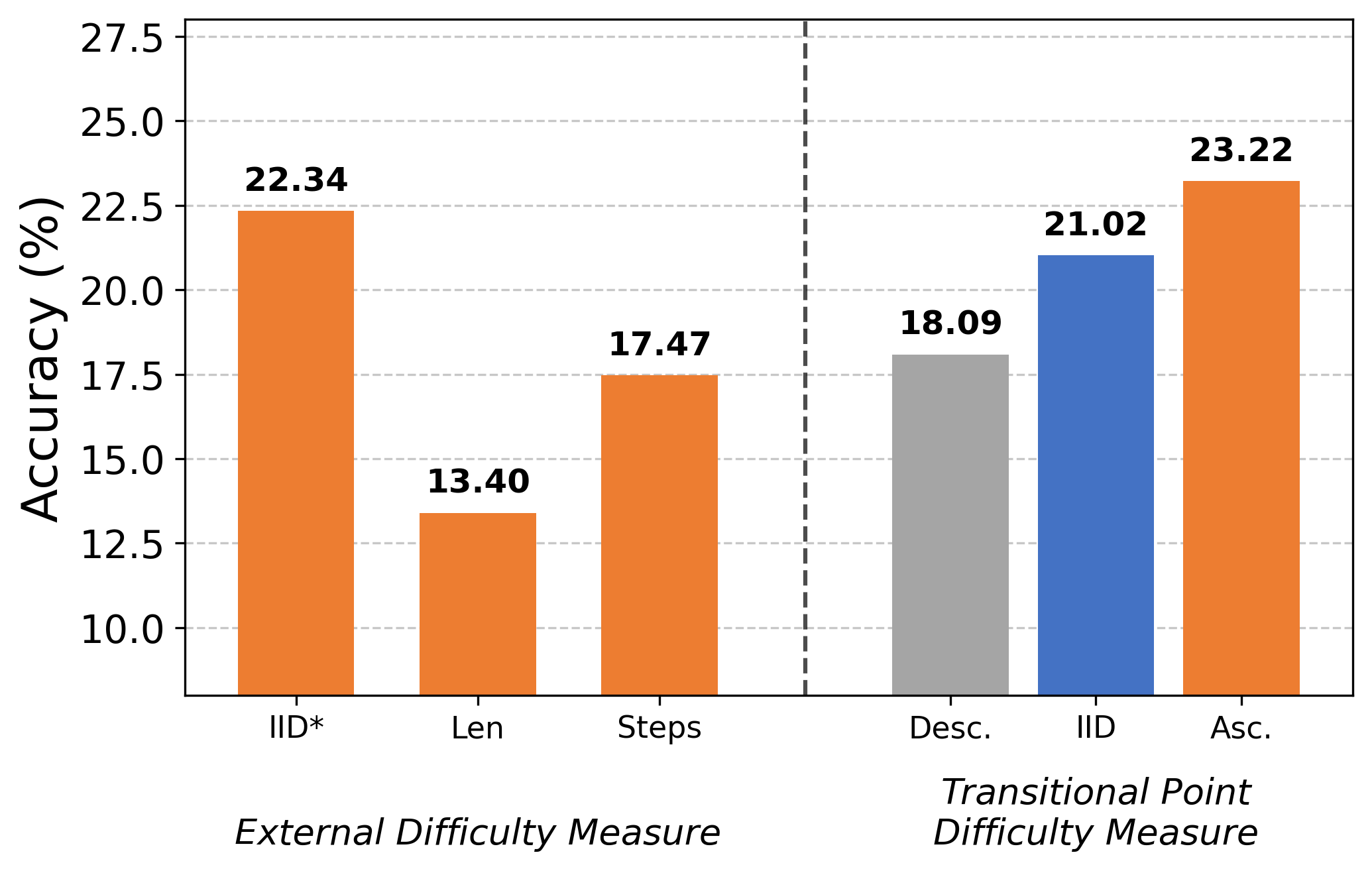}
        \caption{Qwen2.5-0.5B trained on $k$-shot transitional problems from Qwen2.5-1.5B}
        \label{fig:math-curr-kshot-0.5b}
    \end{minipage}%
    \hfill
    \begin{minipage}{0.48\textwidth}
        \centering
        \includegraphics[width=\linewidth]{figures/math_k_shot_main.png}
        \caption{Qwen2.5-1.5B trained on $k$-shot transitional problems from Qwen2.5-1.5B}\label{fig:math-curr-kshot-1.5}
    \end{minipage}%
\end{figure}

For a given series of models ordered by their competence on a task, the set of transitional problems at level $i$ is the subset of problems from that task distribution that are solved by all models that have a competence level $\ge i$ (Section \ref{sec:method}). So far, we have explored two ways to define the model series: via checkpoints from training on the target task distribution, and via a family of models that are naturally increasingly competent due to, e.g., increasing model size. In both of these settings, it is \emph{intuitive} that a stronger model represents a higher tier of competence than a weaker model, and that the weaker model can be improved either by task-specific training or by distillation from stronger models. Now, we examine yet another setup for defining a model series---a model at level $i$ is simply the base (level 0) model to which a test-time few-shot adaptation (such as in-context learning) is applied, with increasing adaptation as the level $i$ increases. Though this setup is less intuitive than the previous settings, it is by far the most efficient to implement, as identifying higher levels does not require the training of the base model or inference with larger models.

To test whether transitional problems defined over the $k$-shot model series are useful to curriculum learning, we use the Qwen2.5-1.5B base model with $k=\{1,2,4,8,16\}$ shots of adaptation. We choose the 1.5 billion parameter model as it shows a consistent increase in performance as additional in-context examples are added, going from $35.8\%$ 0-shot validation accuracy to $73.4\%$ with 16 in-context examples. We then identify the transitional problems from this model series, and train the base Qwen2.5-0.5B and Qwen2.5-1.5B models on the level-up, random, and level-down curricula, in addition to our baseline curricula for the same number of training steps. As seen in Figures \ref{fig:math-curr-kshot-0.5b} and \ref{fig:math-curr-kshot-1.5}, the easy-to-hard curriculum on transitional problems outperforms all other curricula, especially when transferring to the weaker 0.5 billion parameter model.

\begin{figure}[h]
\centering
\includegraphics[width=\linewidth]{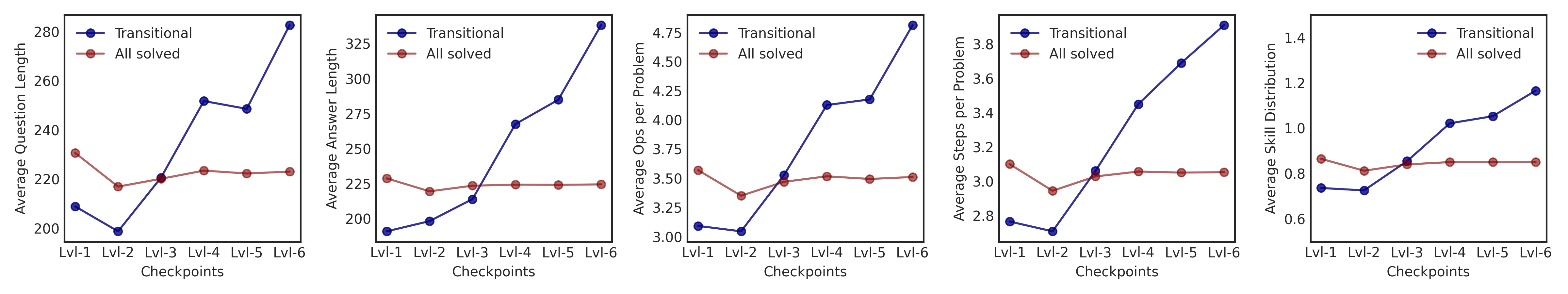}
\caption{Variation in human measures of difficulty for a 6-model series obtained from increasing $k$-shot adaptation on the Qwen2.5-1.5B model. Even with the reduced differentiation in performance between model levels, the transitional problems exhibit a human-interpretable difficulty increase.}\label{fig:math-interp-kshot_5}
\end{figure}

\xhdr{Interpretability} Figure \ref{fig:math-interp-kshot_5} shows that transitional problems corresponding to the $k$-shot transitional series display the same characteristics as the corresponding problems defined in other ways, with some limited jitter. In particular, the differentiation in the average number of operations or the proportion of skills varies more sharply than with other methods, indicating that \emph{in-context examples are useful in recapitulating arithmetic skill} in the Qwen2.5-1.5B model.

\section{Reproducibility}\label{apx:repro}

In this section, we describe our experimental settings in full. We aim to provide enough detail to independently reproduce our results from this paper.

\subsection{Chess Experiments}

\begin{table}[h]
\centering
\caption{Hyperparameter Settings in Chess Experiments.}
\label{tab:hyperparameters_chess}
\resizebox{\textwidth}{!}{
\begin{tabular}{lccccccc}
\toprule
\textbf{Budget} & \textbf{Learning Rate} & \textbf{Weight Decay} & \textbf{\#Train} & \textbf{\#Test} & \textbf{\#Runs} & \textbf{Max Steps} & \textbf{Batch Size} \\
\midrule
Tiny & $10^{-3}$ & $10^{-6}$ & 10,000 & 10,000 & 10 & 90 & 64 \\
Low & $10^{-3}$ & $10^{-6}$ & 10,000 & 10,000 & 10 & 90 & 128 \\
Mid & $10^{-3}$ & $10^{-6}$ & 10,000 & 10,000 & 10 & 180 & 256 \\
\bottomrule
\end{tabular}
}
\end{table}

\subsection{Math Experiments}\label{apx:math_details}

\xhdr{Training checkpoints for the model series} To collect the checkpoints for the model series, we train our base learners (0.5/1.5 billion parameter Qwen2.5 models) on the GSM8k training dataset for 100 steps with an effective batch size of 32 (1.5B) or 64 (0.5B) training samples distributed across 4 NVIDIA A100 GPUs\footnote{\url{https://www.nvidia.com/en-us/data-center/a100/}} and a learning rate of $1\times 10^{-5}$, and collect checkpoints every 5 training steps. We then evaluate the validation accuracy for each checkpoint, and select checkpoints that differ in validation performance by an evenly-spaced interval corresponding to the desired number of models in the series. In this and every other evaluation, we compute the performance of a model as its expected 0-shot `exact match' accuracy to the GSM8k final answer over 8 attempts for every problem (Avg@8), averaged over the evaluated dataset. Following the recommendations of the Qwen team in existing work \citep{jiang2025vcrlvariancebasedcurriculumreinforcement}, we use a temperature of 0.75 and a top-p value of 0.95. We use a generation prompt that is similar to the evaluation prompt in GSM-Symbolic \citep{mirzadeh2025gsmsymbolicunderstandinglimitationsmathematical}.

\xhdr{Qwen2.5 model family series} We evaluate each model in the Qwen2.5 model family as above, but in a 1-shot in-context learning setting (i.e., with 1 training problem and solution as an example in the prompt). This is to ensure that each model formats the answers correctly---the larger Qwen models tend to put the answer between the \verb|\boxed| \LaTeX block, which is incompatible with our answer extraction process. The validation accuracy of the model family ranges from 38\% for the smallest model to over 93\% for the largest.

\xhdr{Transfer learning} Matching Orca problems to GSM8k problems based on learning difficulty is a complex process that we detail here. The Orca dataset is large, but meant for pretraining prior to fine-tuning on GSM8k, and this contains poorly formatted LLM-generated solutions. We use the DeepSeek-R1 distilled Qwen3-8B model \citep{deepseekai2025deepseekr1incentivizingreasoningcapability} in thinking mode to re-format the solutions of 50,000 Orca problems into the style of GSM8k for better problem matching. We use a prompt that encourages mathematical consistency with the original answer, but formatting consistency with 16 provided GSM8k examples. After filtering to ensure the above consistency creiteria are met, our \verb|orca-gsm8k-formatted| dataset consists of roughly 6000 training problems, of which 3800 are training problems, and 2554 are test problems. We then use the Qwen3-Embedding-8B contextual embedding model \citep{zhang2025qwen3embeddingadvancingtext} to embed each GSM8k transitional problem and Orca example into a 256-dimensional vector, with a prompt to the embedding model that encourages the problems to have a high cosine similarity if they are equally difficult. Using manually defined thresholds on cosine similarity, ($\text{mean similarity}\ge 0.65, \text{max similarity}\ge 0.8$), we identify the Orca problems that best correspond to each set of GSM8k transitional problems to produce the \emph{neo-transitional} problems on \verb|orca-gsm8k-formatted|. We then conduct curriculum learning experiments on \verb|orca-gsm8k-formatted| with our model series as in Section \ref{expt:math-curr}.

\xhdr{Curriculum Comparisons}

To collect data for each baseline curriculum (len, steps), we sort the dataset by the answer/question length or steps, respectively, and use a \emph{stratified sampling} strategy to find a subset that spans the full range of lengths (or steps) in the dataset. We repeat each experiment 5 times with varying random seeds for sampling i.i.d. and between each curriculum level, and present averaged results over experiments. We train with the same hyperparameters detailed above. The number of steps in each curriculum varies depending on the number of transitional problems we identify in the GSM8k training set, and is listed in Table \ref{tab:hyperparameters_math} for each setting. Note that each step involves training on 32 (1.5 billion) or 64 (0.5 billion) GSM8k training examples, and our experiments are primarily limited by the training set size of GSM8k and the transitional problems we identify.

\begin{table}[h]
\centering
\caption{Hyperparameter Settings in the Math Experiments.}
\label{tab:hyperparameters_math}
\resizebox{0.9\textwidth}{!}{
\begin{tabular}{lccccccc}
\toprule
\textbf{Size} & \textbf{0.5B-ckpts} & \textbf{1.5B-ckpts} & \textbf{0.5B-family} & \textbf{1.5B-family} & \textbf{0.5B-k-shot} & \textbf{1.5B-k-shot} & \textbf{0.5B-Orca} \\
\midrule
Tiny & 8  & 8 & -- & -- & -- & -- & -- \\
Low & 10 & 10  & 10 & 10 & 10 & 10 & 10\\
Mid & 15 & 12 & -- & 12 & -- & -- & -- \\
\bottomrule
\end{tabular}
}
\end{table}

\begin{table}[htbp]
\centering
\caption{Dataset statistics. \#Source is the total pool used to identify transitional problems; \#Transitional is the number kept for training (identified for Math, capped for Chess); \#Per level lists the per-stage split.}
\label{tab:stats}
\footnotesize
\setlength{\tabcolsep}{4pt}
\begin{tabular}{llllrrl}
\toprule
Setting & Method            & Model              & Dataset         & \#Source     & \#Transitional & \#Per level \\
\midrule
Math    & In-distribution   & Qwen2.5-1.5B-Base  & GSM8k           & 6{,}500      & 2{,}092        & 675 / 462 / 272 / 289 / 394 \\
Math    & Distillation      & Qwen2.5-0.5B-Base  & GSM8k           & 6{,}500      & 5{,}442        & 2{,}181 / 2{,}113 / 693 / 255 \\
Math    & Neo-transitional  & Qwen2.5-0.5B-Base  & Orca-formatted  & 5{,}000      & 539            & 174 / 139 / 114 / 56 / 56 \\
Chess   & In-distribution   & Maia-2             & Puzzles         & $\sim$5.8M   & 180{,}000      & 20{,}000 \\
Chess   & Transfer          & Maia-2             & Game Positions  & $\sim$300B   & 180{,}000      & 20{,}000 \\
\bottomrule
\end{tabular}
\end{table}

\begin{table}[htbp]
\centering
\caption{Per-phase runtime for the math experiments.}
\label{tab:runtime}
\small
\begin{tabular}{lllr}
\toprule
Dataset            & Phase                                 & Model              & Runtime \\
\midrule
GSM8k              & Collect checkpoints                   & Qwen2.5-1.5B-Base  & 25 min 22 sec \\
Transitional/GSM8k & Curriculum (per method/baseline)      & Qwen2.5-1.5B-Base  & 8 min 5 sec \\
\bottomrule
\end{tabular}
\end{table}

%%%%%%%%%%%%%%%%%%%%%%%%%%%%%%%%%%%%%%%%%%%%%%%%%%%%%%%%%%%%

% \newpage
% \clearpage
% \input{checklist.tex}

\end{document}